\pdfoutput=1
\documentclass[11pt]{article}

\usepackage{acl}

\usepackage{times}
\usepackage{latexsym}
\usepackage{enumitem}
\usepackage{booktabs}
\usepackage{multirow}
\usepackage{array}
\usepackage{graphicx}
\usepackage{amsmath}
\usepackage{amsfonts}
\usepackage{subcaption}
\usepackage{makecell}
\usepackage{color,soul}
\usepackage{algorithm}
\usepackage[noend]{algpseudocode}
\usepackage{xcolor,colortbl}
\usepackage{xspace}

\usepackage{url}
\usepackage{hyperref}

\definecolor{green}{rgb}{0.1,0.1,0.1}
\definecolor{gitgreen}{HTML}{006400}
\definecolor{chocolate}{HTML}{D2691E}
\definecolor{maroon}{HTML}{A00000}
\definecolor{indigo}{HTML}{4B0082}
\definecolor{green}{HTML}{008000}

\definecolor{mygold}{HTML}{fdb462}
\definecolor{myred}{HTML}{fb8072}
\definecolor{mypurple}{HTML}{bebada}
\definecolor{mydarkpurple}{HTML}{7570b3}
\definecolor{mygreen}{HTML}{8dd3c7}
\definecolor{mydarkgreen}{HTML}{1b9e77}
\definecolor{myblue}{HTML}{80b1d3}
\newcommand{\goldbox}{\textcolor{mygold}{$\blacksquare$}}
\newcommand{\redbox}{\textcolor{myred}{$\blacksquare$}}
\newcommand{\purplebox}{\textcolor{mypurple}{$\blacksquare$}}
\newcommand{\darkpurplebox}{\textcolor{mydarkpurple}{$\blacksquare$}}
\newcommand{\greenbox}{\textcolor{mygreen}{$\blacksquare$}}
\newcommand{\darkgreenbox}{\textcolor{mydarkgreen}{$\blacksquare$}}
\newcommand{\bluebox}{\textcolor{myblue}{$\blacksquare$}}

\usepackage{amssymb}% http://ctan.org/pkg/amssymb
\usepackage{pifont}% http://ctan.org/pkg/pifont
\newcommand{\cmark}{{\protect\color{maroon} \ding{51}}}
\newcommand{\xmark}{\ding{55}}

\newcommand{\Demo}{Demonstrations}
\newcommand{\demo}{demonstrations}
\newcommand{\gt}{ground truth}

\usepackage[T1]{fontenc}
\usepackage[utf8]{inputenc}

\usepackage{microtype}

\newcommand{\myskip}[1]{}

\title{Rethinking the Role of \Demo: \\
What Makes In-Context Learning Work?}

\newcommand{\affilsup}[1]{\rlap{\textsuperscript{\normalfont#1}}}
\author{
    Sewon Min\affilsup{1,2} \qquad
    Xinxi Lyu\affilsup{1} \qquad
    Ari Holtzman\affilsup{1} \qquad
    Mikel Artetxe\affilsup{2}\\
    \textbf{Mike Lewis}\affilsup{2} \qquad
    \textbf{Hannaneh Hajishirzi}\affilsup{1,3} \qquad
    \textbf{Luke Zettlemoyer}\affilsup{1,2} \\
    $^1$University of Washington \qquad
    $^2$Meta AI \qquad
    $^3$Allen Institute for AI \\
    \texttt{\{sewon,alrope,ahai,hannaneh,lsz\}@cs.washington.edu} \\
    \texttt{\{artetxe,mikelewis\}@meta.com}
}

\begin{document}
\maketitle
\begin{abstract}
%\sewon{I updated the abstract with most comments addressed. One issue is that we are only emphasizing Section 4, and diminishing Section 5.}
Large language models (LMs) are able to in-context learn---perform a new task via inference alone by conditioning on a few input-label pairs (\demo) and making predictions for new inputs.
However, there has been little understanding of {\em how} the model learns and {\em which} aspects of the \demo\ contribute to end task performance.
%In this paper, we show that \gt\ \demo\ are in fact not required---randomly replacing labels in the \demo\ barely hurts performance, consistently over 12 different models including GPT-3.
In this paper, we show that \gt\ \demo\ are in fact not required---randomly replacing labels in the \demo\ barely hurts performance on a range of classification and multi-choce tasks, consistently over 12 different models including GPT-3.
%This indicates that the groundtruth input-label mapping in the \demo\ plays little role in performing the task.
Instead, we find that other aspects of the \demo\ are the key drivers of end task performance, including the fact that they provide a few examples of  
(1) the label space, (2) the distribution of the input text, and (3) the overall format of the sequence.
%Finally, we boost zero-shot performance by using in-domain text and predictions from the LM to create synthetic silver demonstrations. This method outperforms previous zero-shot baseline by 16\%, and archives nearly k-shot performance with gold train data, despite not using any labeled data.
Together, our analysis provides a new way of understanding how and why in-context learning works, while opening up new questions about how much can be learned from large language models through inference alone. 
\end{abstract}

\section{Introduction}\label{sec:intro}Large language models (LMs) have shown impressive performance on downstream tasks by simply conditioning on a few input-label pairs (\demo); this type of inference has been referred to as {\em in-context learning}~\citep{brown2020language}.
%are capable of {\em in-context learning}---they can perform a new task during inference by simply conditioning on a few input-label pairs (called {\em \demo})~\citep{brown2020language}. % to make a prediction to the test input.
%Prior work has suggested that the model {\em ``learns''} a new task from these input-label pairs as humans do~\citep{brown2020language,zhao2021calibrate,liu2021makes}.
Despite in-context learning consistently outperforming zero-shot inference on a wide range of tasks~\citep{zhao2021calibrate,liu2021makes}, 
there is little understanding of {\em how} it works and {\em which} aspects of the \demo\ contribute to end task performance.

\begin{figure}[t]
\centering \footnotesize
%\resizebox{\columnwidth}{!}{\includegraphics[trim={11cm 18.5cm 6cm 2cm},clip]{figures/composition.pdf}}
\resizebox{\columnwidth}{!}{\includegraphics[trim={4cm 0.5cm 4cm 0.8cm},clip]{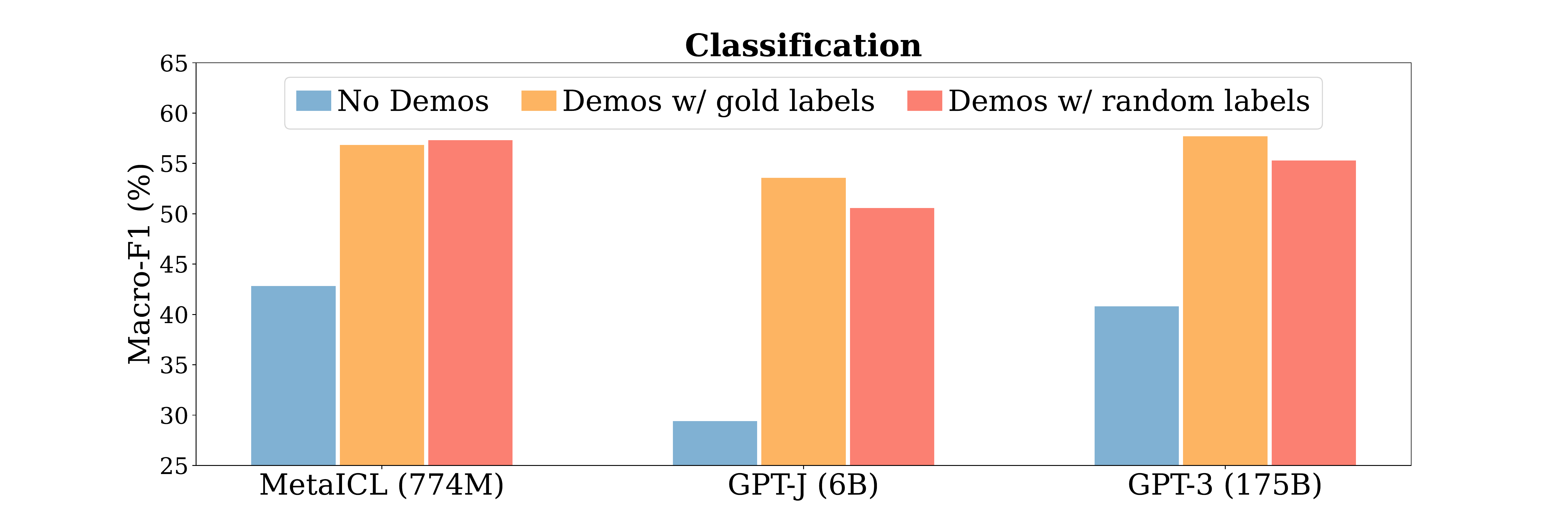}}
\resizebox{\columnwidth}{!}{\includegraphics[trim={4cm 0.5cm 4cm 0.8cm},clip]{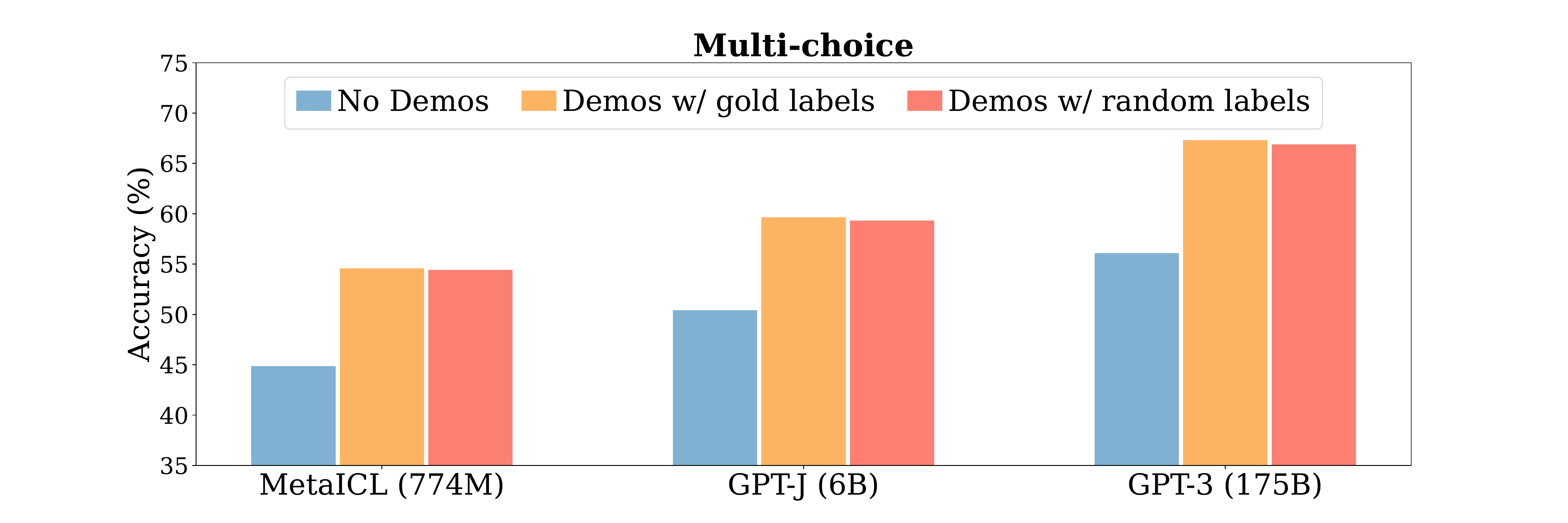}}
\caption{
    Results in classification (top) and multi-choice tasks (bottom), using three LMs with varying size.
    %with 774M, 6B, and 175B parameters, respectively,
    %\mikel{you can report model sizes in the figure to make the caption more lightweight, ie MetaICL (774M), GPT-J (6B), GPT-3 (175B) in the X axis}
    %(all with channel).
    %\ari{most people won't know what "with channel" means here until they read the rest of the paper, and it's not really important for the immediate point. I would just include a reference to the experimental section that describes the details instead of trying to fill them in here. Probably the same for the next sentence, sinc e you want to get to  the last sentence as quickly as possible.}
    %\sewon{I think some people might still want to see some clarification about the experiment.. I squeezed details into one sentence---is it better?}
    Reported on six datasets on which GPT-3 is evaluated; %\mikel{similar to above, maybe report exact tasks in parenthesis to make the caption shorter, eg Classification (x, y, z)};
    the channel method is used.
    %\mikel{any reason to use the channel approach here? direct seems more standard and easier to justify}
    See Section~\ref{sec:main} for the full results.
    %\mikel{larger-scale evaluation -> full results?}
    % on 26 datasets and with 12 models.
    %Reported on six datasets on which GPT-3 is evaluated; full results on 6 datasets for other models are in Figure~\ref{fig:main-results}.
    In-context learning performance drops only marginally when labels in the \demo\ are replaced by random labels.
    %\mikel{No Demos -> Zero-shot, Demos -> Few-shot seems more standard}
    %\ari{ideally, you want this graphic to make sense before someone reads the caption. therefore, I recommend the labels "No Demo", "Gold Labels", and "Random Labels"—it's slightly less specific, but just saying "random" without a noun is too confusing and ruins the "instant communication" factor.}
    %\ari{just leaving a little note here that the diagram should eventually label the top and bottom parts of the figure visually not just in the caption.}
}\label{fig:intro}
\end{figure}

In this paper, we show that \gt\ \demo\ are in fact not required for effective in-context learning (Section~\ref{sec:main}).
%Specifically, replacing the labels in \demo\ with random labels barely hurts performance (Figure~\ref{fig:intro}).
Specifically, replacing the labels in \demo\ with random labels barely hurts performance in a range of classification and multi-choice tasks (Figure~\ref{fig:intro}).
% correctly.
%Model performance only drops marginally when randomly replacing the labels in the \demo\ (Figure~\ref{fig:intro}).
%that the input-label correlation in the \demo\ in fact matter significantly less than previously thought (Section~\ref{sec:main}).
%Model performance drops only marginally when labels in the \demo\ are randomly replaced (Figure~\ref{fig:intro}).
The result is consistent over 12 different models including the GPT-3 family~\citep{radford2019language,min2021metaicl,wang2021gpt,artetxe2021efficient,brown2020language}.
%It, overall, strongly suggests that there must be other aspects of the \demo, beyond the mapping from inputs to outputs, that are allowing the model to perform the task. 
%This strongly suggests that the model is not using the \demo\ to learn the {\em correspondence} between inputs and labels required to perform the downstream task.
This strongly suggests, counter-intuitively, that the model {\em does not} %require the groundtruth input-label mapping 
rely on the input-label mapping in the \demo\
to perform the task.

Further analysis investigates which parts of \demo\ actually {\em do} contribute to the performance.
We identify possible aspects of \demo\
(e.g., the label space and the distribution of the input text)
and evaluate a series of variants of the \demo\ to quantify the impact of each (Section~\ref{sec:abl}).
%
%We identify three aspects of \demo\ aside the input-label mapping---the {\em label space}, the { \em distribution of inputs}, and {\em the format of having the input-label pairs}---and vary the composition of the \demo\ to quantify the impact of each (Section~\ref{sec:abl}).
%
We find that: (1) the label space and the distribution of the input text {\em specified by} the \demo\ are both key to in-context learning (regardless of whether the labels are correct for individual inputs);
%{ \em independent } of whether inputs and labels are correctly matched;
%(2) In the case of classification, using random English words as labels significantly outperforms only using input examples (no labels), indicating the importance of the { \em input-label pair } format.
(2) specifying the overall
%examples
format is also crucial, e.g.,
%simply keeping the format of the \demo\ with random English words as labels achieves substantial performance gains;
when the label space is unknown, using random English words as labels is significantly better than using no labels;
and
% (3) if models are finetuned with an in-context learning objective~\citep{min2021metaicl}, models are encouraged to exploit simpler aspects of the \demo\ like the format, rather than the groundtruth input-label mapping.
(3) meta-training with an in-context learning objective~\citep{min2021metaicl} 
magnifies these effects---the models almost exclusively exploit simpler aspects of the \demo\ like the format rather than the input-label mapping.

In summary, our analysis provides a new way of understanding the role of the \demo\ in in-context learning.
We empirically show that the model (1) counter-intuitively does not rely on the \gt\ input-label mapping provided in the \demo\ as much as we thought (Section~\ref{sec:main}), and (2) nonetheless still benefits from knowing the label space and the distribution of inputs specified by the \demo\ (Section~\ref{sec:abl}).
We also include a discussion of broader implications, e.g., what we can say about the model {\em learning at test time}, and avenues for future work (Section~\ref{sec:discuss}).

%To summarize, our contributions are three-fold:
%\begin{enumerate}[itemsep=0em]
%    \item We find that the groundtruth input-label mapping in the \demo\ plays only a small role in in-context learning, consistently over 12 different models (Section~\ref{sec:main}).
%    \item We empirically quantify other aspects of the \demo\ that contribute to performance: the label space, the distribution of inputs, and the use of input-label pairs (Section~\ref{sec:abl}).
%    \item We include discussion of broader implications and avenues for future work (Section~\ref{sec:discuss}).
%\end{enumerate}

\section{Related Work}\label{sec:related}
\begin{figure}[!t]
\centering \footnotesize
\resizebox{\columnwidth}{!}{\includegraphics[trim={16cm 15cm 12cm 7cm},clip]{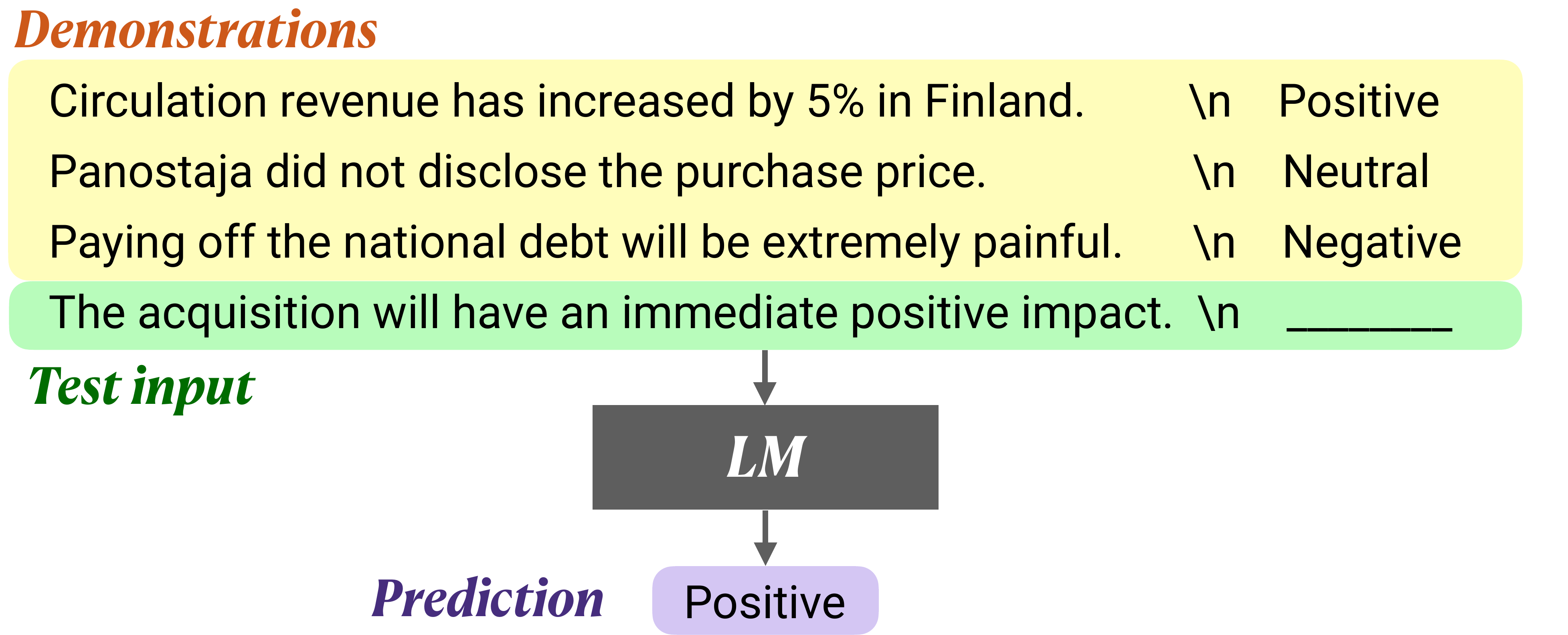}}
\caption{
    An overview of in-context learning. The demonstrations consist of $k$ input-label pairs from the training data ($k=3$ in the figure).
}\label{fig:icl}
\end{figure}

Large language models have been key to strong performance
%improvements
in a wide range of downstream tasks~\citep{devlin2019bert,radford2019language,liu2019roberta,raffel2020exploring,lewis2020bart}. While finetuning has been a popular approach to transfer to new tasks~\citep{devlin2019bert}, it is often impractical to finetune a very large model (e.g. $\geq$10B parameters).
\citet{brown2020language} propose in-context learning as an alternative way to learn a new task. As depicted in Figure~\ref{fig:icl}, the LM learns a new task via inference alone by conditioning on a concatenation of the training data as \demo, without any gradient updates.

In-context learning has been the focus of significant study since its introduction. Prior work proposes
%a better way of prompting the model~\citep{shin2020autoprompt,gao2021making,li2021prefix,lester2021power},
better ways of formulating the problem~\citep{zhao2021calibrate,holtzman2021surface,min2021noisy}, better ways of choosing labeled examples for the \demo~\citep{liu2021makes,lu2021fantastically,rubin2021learning}, meta-training with an explicit in-context learning objective~\citep{chen2021meta,min2021metaicl}, and learning to follow instructions as a variant of in-context learning~\citep{mishra2021cross,efrat2020turking,wei2022finetuned,sanh2022multitask}.
At the same time, some work reports brittleness and over-sensitivity  for in-context learning~\citep{lu2021fantastically,zhao2021calibrate,mishra2021reframing}.

Relatively less work has been done to understand why in-context learning works.
\citet{xie2022explanation} provide theoretical analysis that in-context learning can be formalized as Bayesian inference that uses the \demo\ to recover latent concepts.
\citet{razeghi2022impact} show that in-context learning performance is highly correlated with term frequencies in the pretraining data.
%\citet{reynolds2021prompt} claim the model already has the capacity to perform the task without \demo\ when the right prompts are used.
To the best of our knowledge, this paper is the first that provides an empirical analysis that investigates why in-context learning achieves performance gains over zero-shot inference.
We find that the \gt\ input-label mapping in the \demo\ has only a marginal effect, and measure the impact of finer-grained aspects of the \demo.

\begin{table}[t]
    \centering \footnotesize
    \begin{tabular}{l @{\hspace{2em}} rrr}
        \toprule
            Model & \# Params & Public & Meta-trained \\
        \midrule
            GPT-2 Large & 774M  & \cmark & \xmark \\
            MetaICL         & 774M  & \cmark & \cmark \\
            GPT-J               & 6B    & \cmark & \xmark \\
            %MetaICL 6B~\citep{min2021metaicl}       & 6B    & \cmark & \cmark \\
            fairseq 6.7B$^\dagger$ & 6.7B & \cmark & \xmark \\
            fairseq 13B$^\dagger$ & 13B   & \cmark & \xmark \\
            GPT-3      & 175B$^\ddagger$  & \xmark & \xmark \\
        \bottomrule
    \end{tabular}\vspace{-.1em}
    \caption{
        A list of LMs used in the experiments:
        GPT-2~\citep{radford2019language},
        MetaICL~\citep{min2021metaicl}, GPT-J~\citep{wang2021gpt}, fairseq LMs~\citep{artetxe2021efficient} and GPT-3~\citep{brown2020language}.
        `Public' indicates whether the model weights are public;
        %\mikel{i'd say model weights are public, as some might argue gpt3 is also public through its api}
        `Meta-trained' indicates whether the model is meta-trained with an in-context learning objective.
        $^\dagger$We use dense models in \citet{artetxe2021efficient} and refer them as fairseq LMs for convenience.
        $^\ddagger$We use the Davinci API (the {\em base} version, not the {\em instruct} version) and assume it to be 175B, following~\citet{gao2021pile} and \citet{artetxe2021efficient}.
        %\mikel{might be worth to clarify that we don't use the instructable models, which I think are the default now}
        %\ari{I would strongly encourage changing "trained" to "finetuned", because currently this is going to cause immediate alarm-bells to go up when most people read this sentence.}
    }\label{tab:models}
\end{table}

\section{Experimental Setup}\label{sec:setup}\begin{figure*}[!t]
\centering \footnotesize
\resizebox{2.1\columnwidth}{!}{\includegraphics[trim={0 0.5cm 0 0.5cm},clip]{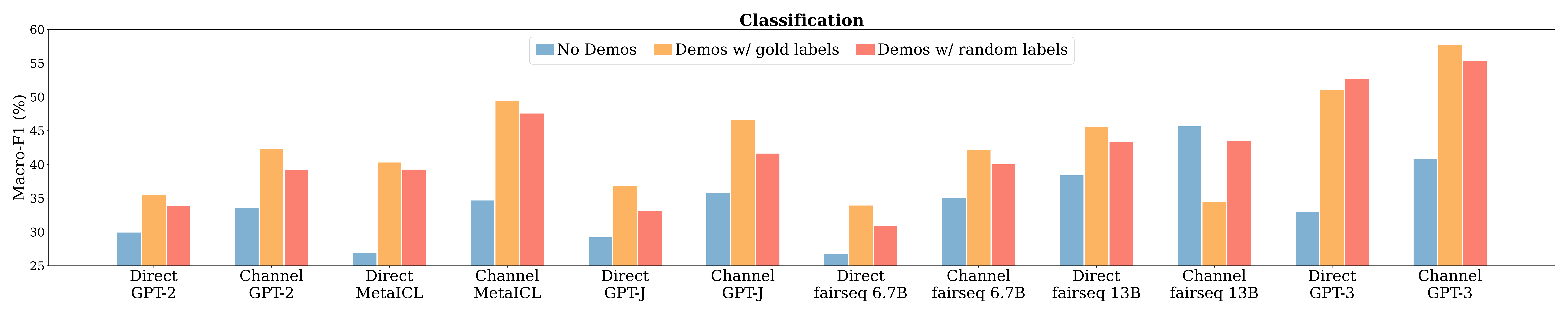}}
\resizebox{2.1\columnwidth}{!}{\includegraphics[trim={0 0.5cm 0 0.5cm},clip]{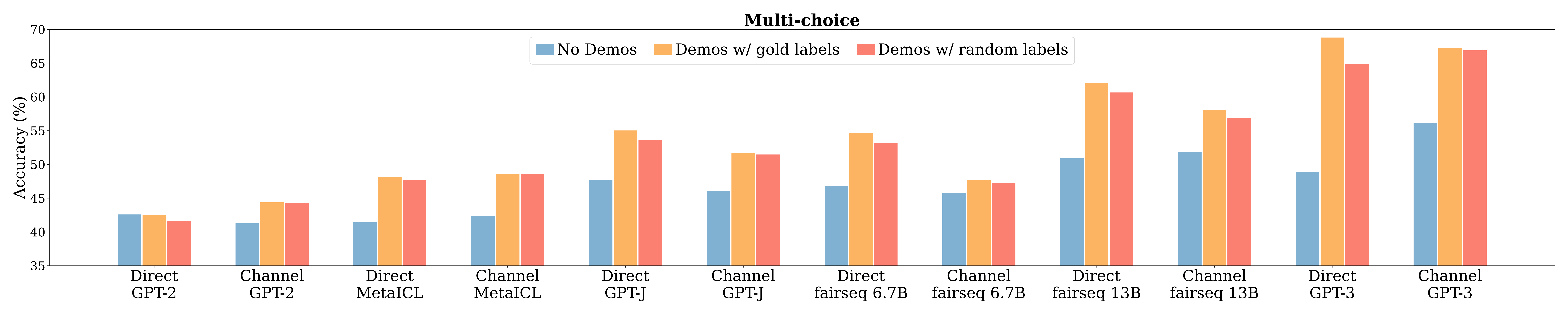}}
\caption{
    Results when using no-\demo, \demo\ with gold labels, and \demo\ with random labels in classification (top) and multi-choice tasks (bottom).
    The first eight models are evaluated on 16 classification and 10 multi-choice datasets, and the last four models are evaluated on 3 classification and 3 multi-choice datasets. See Figure~\ref{fig:app-main-results} for numbers comparable across all models.
    \textbf{Model performance with random labels is very close to performance with gold labels} (more discussion in Section~\ref{subsec:main}). %\ari{You don't need a ``-'' in ``No Demos''}
}\label{fig:main-results}
\end{figure*}

We describe the experimental setup used in our analysis (Section~\ref{sec:main} and \ref{sec:abl}).

\vspace{-.1em}
\paragraph{Models.}
We experiment with 12 models in total.
We include 6 language models (Table~\ref{tab:models}), all of which are decoder-only, dense LMs.
%We use each LM with two inference methods, direct and channel, which compute the conditional probability of the label given the input, and the conditional probability of the input given the label~\citep{min2021noisy}, respectively.
We use each LM with two inference methods, direct and channel, following \citet{min2021noisy}.
%\mikel{I find this terminology a bit strange. I'd say you compare 12 different systems (and not models) from combining 6 models (and not model classes) and 2 approaches or inference methods.}
The sizes of LMs vary from 774M to 175B.
We include the largest dense LM (GPT-3) %\mikel{you should say the largest dense LM as MoEs are larger, but GPT-3 is not even the largest dense LM anymore.}
and the largest publicly released dense LM (fairseq 13B) %\mikel{no longer true, there is GPT-NeoX-20B now}
at the time of conducting experiments.
%\mikel{maybe just say at the time of conducting these experiments, which leaves more margin?}
We also include MetaICL, which is initialized from GPT-2 Large and then meta-trained on a collection of supervised datasets with an in-context learning objective, and ensure that our evaluation datasets do not overlap with those used at meta-training time.

\vspace{-.1em}
\paragraph{Evaluation Data.}
We evaluate on 26 datasets, including sentiment analysis, paraphrase detection, natural language inference, hate speech detection, question answering, and sentence completion (full list and references provided in Appendix~\ref{app:datasets}).\footnote{
For convenience, we use `labels' to refer to the output for the task, though our datasets include non-classification tasks.}
All datasets are classification and multi-choice tasks.

We use these datasets because they (1) are true low-resource datasets with less than 10K training examples, (2) include well-studied benchmarks from GLUE~\citep{wang2018glue} and SuperGLUE~\citep{wang2019superglue}, and (3) cover diverse domains including science, social media, finance, and more.
%The 26 datasets can be further broken down into 16 classification tasks and 10 multi-choice tasks.

\begin{figure*}[!t]
\centering \footnotesize
\resizebox{2.1\columnwidth}{!}{\includegraphics[trim={8cm 0cm 8cm 1cm},clip]{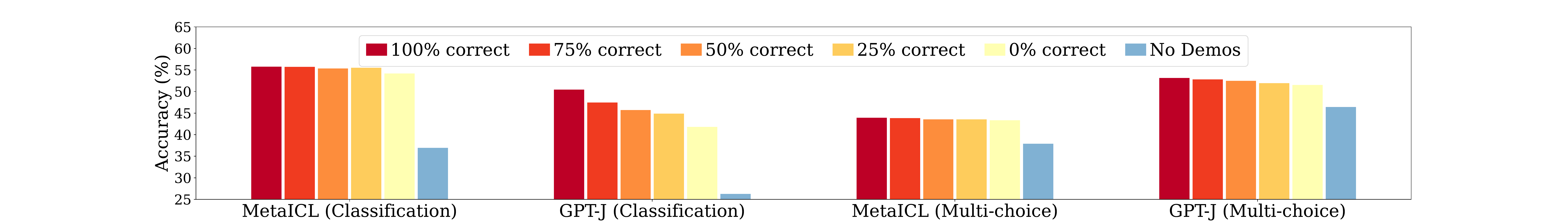}}
\caption{
    Results with varying number of correct labels in the \demo. %MetaICL and GPT-J (both channel) are used.
    Channel and Direct used for classification and multi-choice, respectively.
    Performance with no \demo\ (blue) is reported as a reference.
    %\ari{reminder to change ``No-Demons'' to ``No Demos''.}
}\label{fig:abl_accuracy}
\end{figure*}

\vspace{-.1em}
\paragraph{Other Details.}
We use $k=16$ examples as demonstrations by default for all experiments in the paper, unless otherwise specified.
Examples are sampled at uniform from the training data.
%\mikel{16 demonstrations per class or 16 in total? If it's the former we should clarify it, if it's the latter might be good to further specify 16 demonstrations sampled uniformly from the training set or something around those lines.}
%For all methods that use the demonstrations,
We choose a set of $k$ training examples using 5 different random seeds and run experiments 5 times.
For fairseq 13B and GPT-3, due to limited resources, we experiment with a subset of 6 datasets\footnote{
Three classification and three multi-choice: % datasets:
MRPC, RTE, Tweet\_eval-hate, OpenbookQA, CommonsenseQA, COPA.} and 3 random seeds.
We report Macro-F1\footnote{Known to be better for imbalanced classes.}
for classification tasks and Accuracy for multi-choice tasks.
We compute per-dataset average over seeds, and then report macro-average over datasets.
%We follow the standard for each LM in preprocessing such as separating examples in the \demo.
We use the minimal templates in forming an input sequence from an example. %, e.g., using the label words from the dataset instead of manually-written verbalizers. %\mikel{not obvious what minimal template means here, i think we should further elaborate this and if possible justify the decision.}
We refer to Appendix~\ref{app:exp-details} for more details. All experiments are reproducible from \href{https://github.com/Alrope123/rethinking-demonstrations}{\nolinkurl{github.com/Alrope123/rethinking-demonstrations}}.
\section{Ground Truth Matters Little}\label{sec:main}%\ari{I think pretty much all the subsections would be clearer if you wrote what hypothesis you were testing, instead of what part of the demonstrations you were varying.}
%\sewon{done! I really like this update in Section 5. For Section 4, I think the previous version was OK because Section 4.1 is the main experiments and others are sort of follow-up ablations of the same research question.}
%\sewon{I actually revert it back because we have many ablations.}

%\subsection{How important %are correct labels?
%is groundtruth?
%}\label{subsec:main}
\subsection{Gold labels vs. random labels}\label{subsec:main}
To see the impact of correctly-paired inputs and labels in the \demo---which we call the \gt\ input-label mapping---we compare the following three methods.\footnote{
Without loss of generality, all methods in Section~\ref{sec:main} and \ref{sec:abl} are described based on the direct method, but can be trivially converted to the channel method by flipping $x$ and $y$.}

\vspace{.3em}
\noindent
\textbf{No \demo} is a typical zero-shot method that does not use any labeled data. A prediction is made via $\mathrm{argmax}_{y \in \mathcal{C}}P(y|x)$, where $x$ is the test input and $\mathcal{C}$ is a small discrete set of possible labels.

\vspace{.3em}
\noindent
\textbf{\Demo\ w/ gold labels} are used in a typical in-context learning method with $k$ labeled examples $(x_1, y_1)...(x_k, y_k)$. A concatenation of $k$ input-label pairs is used to make a prediction via $\mathrm{argmax}_{y \in \mathcal{C}}P(y|x_1,y_1...x_k,y_k,x)$.

\vspace{.3em}
\noindent
\textbf{\Demo\ w/ random labels} are formed with random labels, instead of gold labels from the labeled data. Each $x_i$ ($1 \leq i \leq k$) is paired with $\tilde{y}_i$ that is randomly sampled at uniform from $\mathcal{C}$. A concatenation of $(x_1, \tilde{y}_1)...(x_k, \tilde{y}_k)$ is then used to make a prediction via $\mathrm{argmax}_{y \in \mathcal{C}}P(y|x_1,\tilde{y}_1...x_k,\tilde{y}_k,x)$.

\vspace{.5em}
Results are reported in Figure~\ref{fig:main-results}.
First, using the \demo\ with gold labels significantly improves the performance over no \demo,\footnote{
    There are some exceptions, e.g., in the classification tasks, Direct GPT-2, Direct GPT-J and Direct fairseq 6.7B models are not significantly better than random guessing on many datasets; Channel fairseq 13B has significantly better no-\demo\ performance compared to \demo\ with gold labels. We thus discuss the results from these models less significantly for the rest of analysis.
}
as it has been consistently found in much of prior work~\citep{brown2020language,zhao2021calibrate,liu2021makes}.
We then find that \textbf{replacing gold labels with random labels only marginally hurts performance}.
The trend is consistent over nearly all models: models see performance drop in the range of 0--5\% absolute.
%(1.2\% when averaged across all models).
There is less impact in replacing labels in multi-choice tasks (1.7\% on average) than in classification tasks (2.6\% absolute).

\vspace{.2em}
This result indicates that the \gt\ input-label pairs are not necessary to achieve performance gains.
This is counter-intuitive, given that correctly paired training data is critical in typical supervised training---it informs the model of the expected input-label {\em correspondence} required to perform the downstream task.
Nonetheless, the models {\em do} achieve non-trivial performance on the downstream tasks.
This strongly suggests that the models are capable of recovering the expected input-label correspondence for the task; however, it is {\em not} directly from the pairings in the \demo.
%\ari{I think we'll need to expand this description a bit—people who haven't thought deeply about ICL will not find it easy to parse. We need to suggest the natural interpretation, that correspondence matters, then show why it doesn't}

\vspace{.2em}
It is also worth noting that there is particularly little performance drop in MetaICL: 0.1--0.9\% absolute. This suggests that meta-training with an explicit in-context learning objective actually encourages the model to essentially ignore the input-label mapping and exploit other components of the \demo\ (more discussion in Section~\ref{subsec:abl_metaicl}).

In Appendix~\ref{app:task-breakdown}, we provide additional results showing that (1) selecting random labels from a true distribution of labels (instead of a uniform distribution) reduces the gap even further, and (2) the trends may depend on the dataset, although the overall trend is consistent over most datasets.

%In particular, in multi-choice tasks, 72--99\% of improvements through using the \demo\ with gold labels is preserved\footnote{Preserved ratio calculated via $\frac{a_\mathrm{r}-a_\mathrm{n}}{a_\mathrm{g}-a_\mathrm{n}}$, where $a_\mathrm{n}, a_\mathrm{g}, a_\mathrm{r}$ are the accuracy of \noDemos, \gold, \random, respectively.} when the \demo\ with random labels (with an exception in Direct GPT-2 where the \demo\ does not help). In classification tasks, 52--92\% of improvements from the \demo\ is preserved with random labels (with an exception in the channel fairseq 13B model where the \demo\ does not help. The preserved ratio is particularly high (87\%--99\%) in MetaICL.

%Overall, the result suggests that the correct correlation in the demonstration is not necessary to achieve performance gains through in-context learning, and gains from in-context learning does not come from informing the model of the expected relationship between inputs and labels.

\subsection{Ablations}\label{subsec:abl_main}
For additional ablations, we experiment with 5 classification and 4 multi-choice datasets.\footnote{
Classification includes: MRPC, RTE, Tweet\_eval-hate, SICK, poem-sentiment; Multi-choice includes OpenbookQA, CommonsenseQA, COPA and ARC.
}

\begin{figure}[t]
\centering \footnotesize
\resizebox{.49\columnwidth}{!}{\includegraphics[trim={0cm 0cm 2cm 1cm},clip]{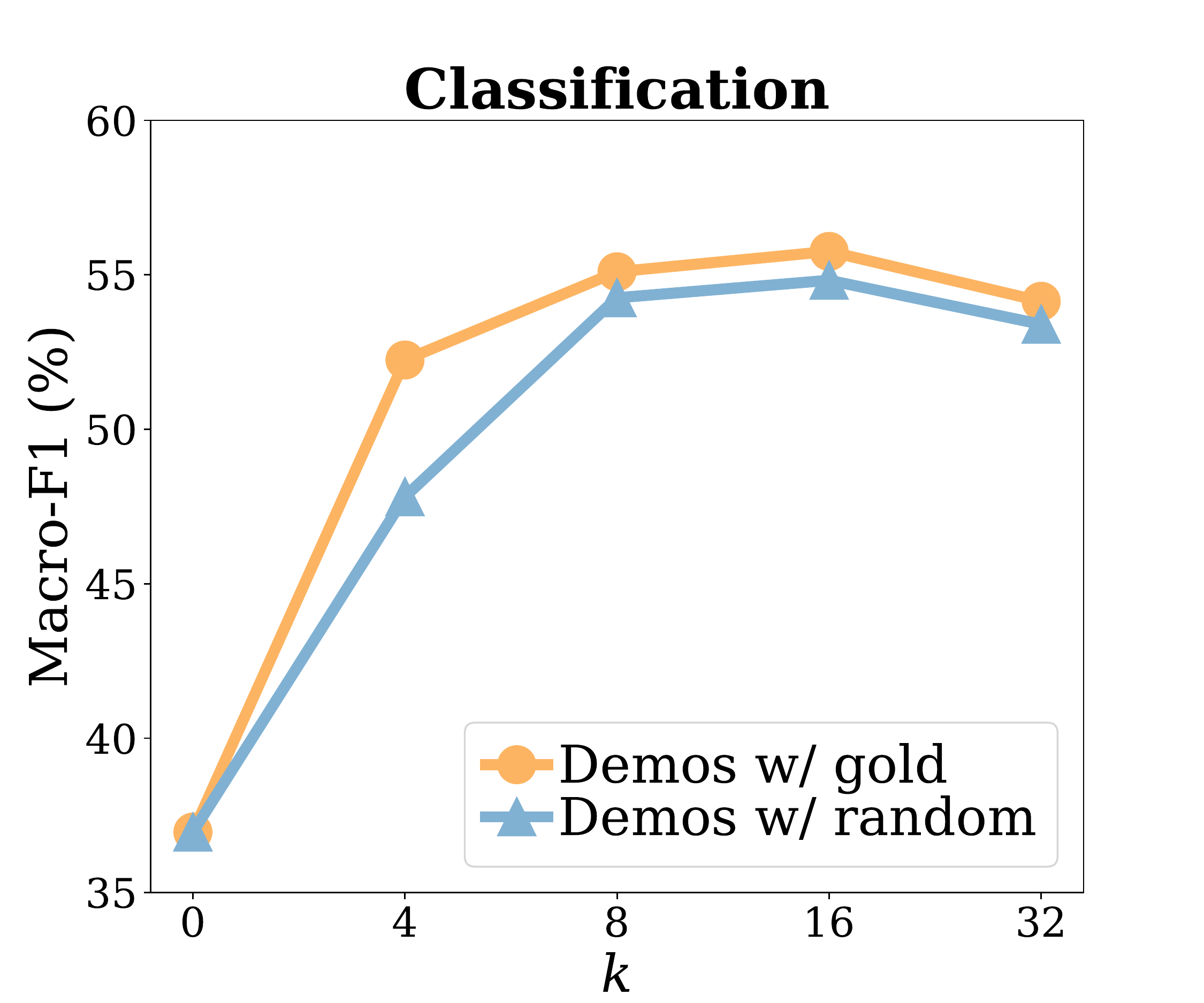}}
\resizebox{.49\columnwidth}{!}{\includegraphics[trim={0cm 0cm 2cm 1cm},clip]{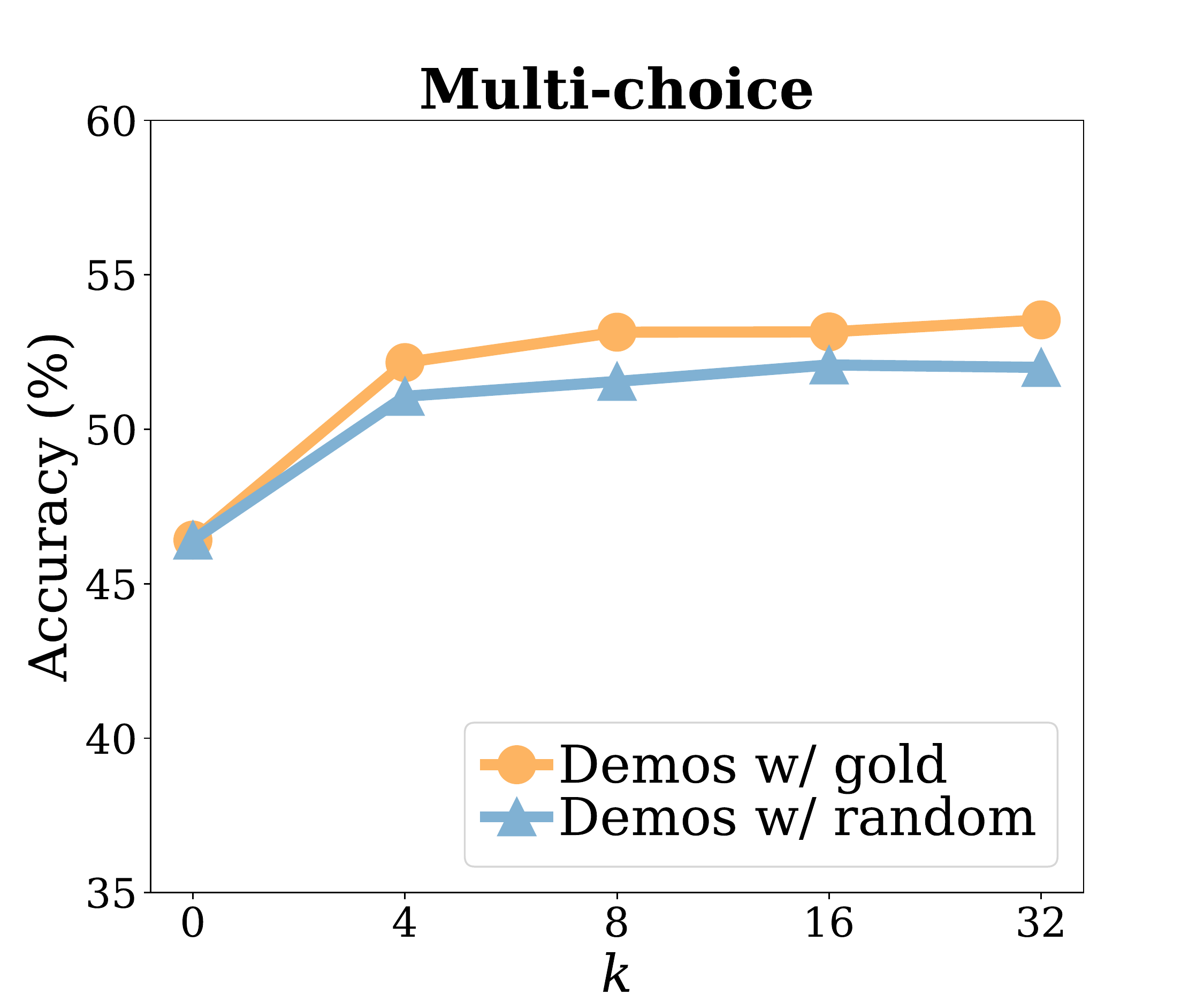}}
\caption{
    Ablations on varying numbers of examples in the \demo\ ($k$). Models that are the best under 13B in each task category (Channel MetaICL and Direct GPT-J, respectively) are used. 
}\label{fig:abl_k}
\end{figure}
\begin{figure*}[!t]
\centering \footnotesize
\resizebox{2.1\columnwidth}{!}{\includegraphics[trim={8cm 0cm 8cm 1cm},clip]{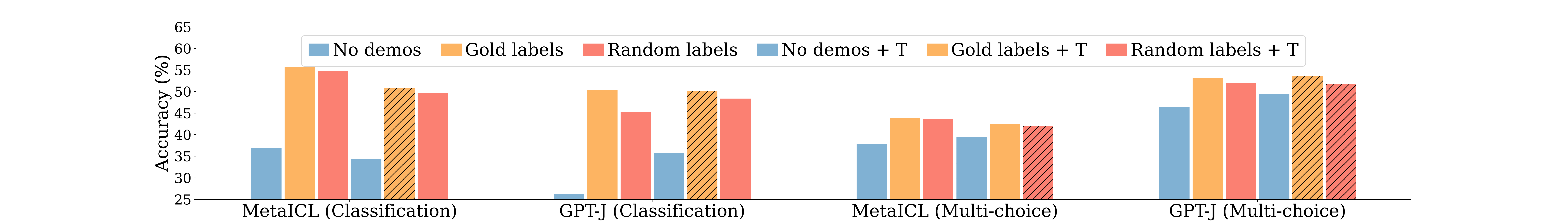}}
\caption{
    Results with minimal templates and manual templates. `+T' indicates that manual templates are used. Channel and Direct used for classification and multi-choice, respectively.
}\label{fig:abl_template}
\end{figure*}

\vspace{-.1em}
%\paragraph{Impact of accuracy in the \demo.}
\paragraph{Does the number of correct labels matter?}
To further examine the impact of correctness of labels in the \demo, we conduct an ablation study by varying the number of correct labels in the \demo.
We evaluate ``\Demo\ w/ $a$\% correct labels'' ($0 \leq a \leq 100$) which consist of $k \times a/100$ correct pairs and $k \times (1-a/100)$ incorrect pairs (see Algorithm~\ref{alg:abl_accuracy} in Appendix~\ref{app:exp-details}).
Here, $a=100$ is the same as typical in-context learning, i.e., \demo\ w/ gold labels.

Results are reported in Figure~\ref{fig:abl_accuracy}.
Model performance is fairly insensitive to the number of correct labels in the \demo.
In fact, always using incorrect labels significantly outperforms no-\demo, e.g., preserving 92\%, 100\% and 97\% of improvements from using the \demo\ with MetaICL in classification, MetaICL in multi-choice, and GPT-J in multi-choice, respectively.
%\ari{It's a bit confusing to compare to no \demo and then use few-shot w/Gold as the comparison point...}
%GPT-J in classification is an outlier where performance depends relatively more on the number of correct labels of the \demo---it achieves higher performance with a larger number of correct labels. Still, always using incorrect labels is significantly better than no \demo.
In contrast, GPT-J in classification sees relatively significant performance drop with more incorrect labels, e.g., nearly 10\% drop in performance when always using incorrect labels. Still, always using incorrect labels is significantly better than no \demo.
%\ari{Hmm...this seems like it just means unsupervised models need to know about the label space for classification, which would actually make perfect sense. I know we're trying to be quick, but it seems worth checkin this with other models?} \sewon{Isn't it basically what experiments with random labels show?}

\vspace{-.1em}
\paragraph{Is the result consistent with varying $\boldsymbol{k}$?}

We study the impact of the number of input-label pairs ($k$) in the \demo. Results are reported in Figure~\ref{fig:abl_k}.
First, using the \demo\ significantly outperforms the no \demo\ method even with small $k$ ($k=4$), and performance drop from using gold labels to using random labels is consistently small across varying $k$, in the range of 0.8--1.6\%.\footnote{With an exception of 4.4\% in classification with $k=4$, likely due to a high variance with a very small value of $k$.}
Interestingly, model performance does not increase much as $k$ increases when $k \geq 8$, both with gold labels and with random labels.
This is in contrast with typical supervised training where model performance rapidly increases as $k$ increases, especially when $k$ is small. We hypothesize that larger labeled data is beneficial mainly for supervising the input-label correspondence, and other components of the data like the example inputs, example labels and the data format are easier to recover from the small data, which is potentially a reason for minimal performance gains from larger $k$ (more discussion in Section~\ref{sec:abl}). %\ari{I like this section a lot! very clear and insightful}

\vspace{-.1em}
\paragraph{Is the result consistent with better templates?}
While we use minimal templates by default, we also explore manual templates, i.e., templates that are manually written in a dataset-specific manner, taken from prior work (details in Appendix~\ref{app:exp-details}). Figure~\ref{fig:abl_template} shows that the trend---replacing gold labels with random labels barely hurting performance---holds with manual templates. It is worth noting that using manual templates does not always outperform using minimal templates.

\section{Why \textit{does} In-Context Learning work?}\label{sec:abl}Section~\ref{sec:main} shows that the \gt\ input-label mapping in the \demo\ has little impact to performance gains from in-context learning.
This section further examines what other aspects of the \demo\ lead to good performance of in-context learning.

\begin{figure}[!t]
\centering \footnotesize
\resizebox{\columnwidth}{!}{\includegraphics[trim={14cm 18.5cm 6.5cm 2cm},clip]{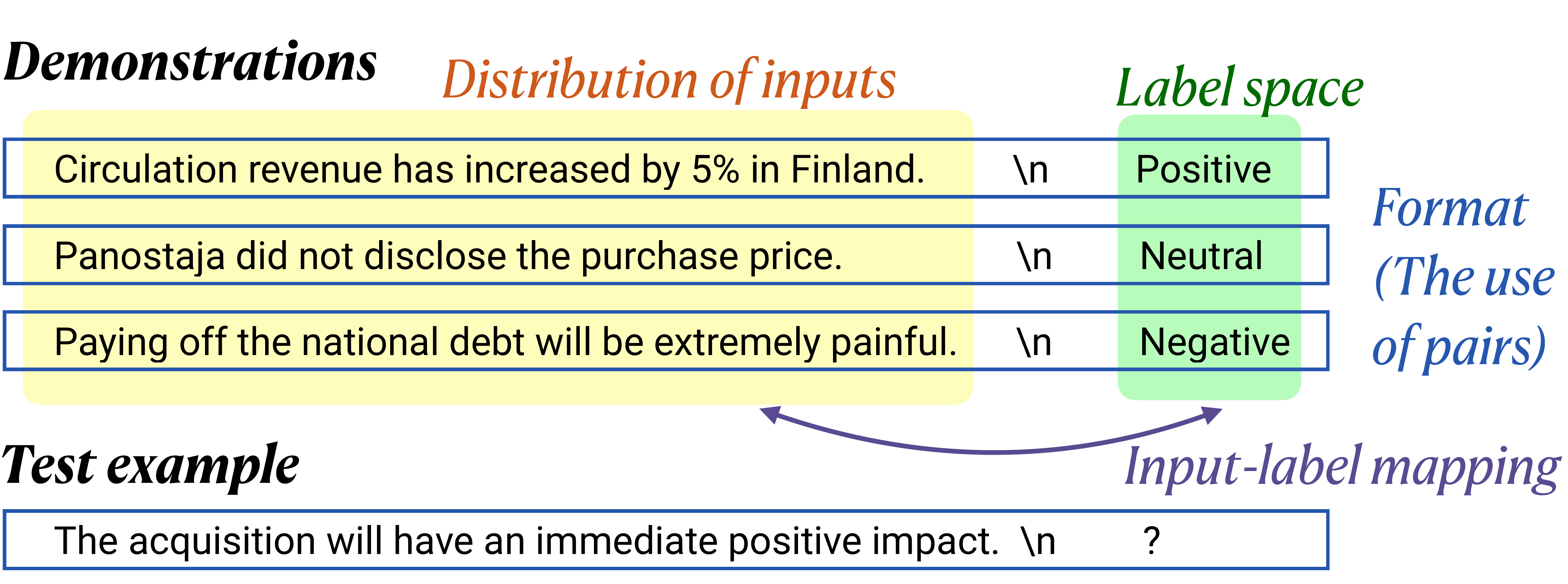}}
\caption{
    Four different aspects in the \demo: the input-label mapping,
    %the example inputs, the example labels and the form of the ``pairs'' (e.g., having the input-label pairs as the {\em form} of the \demo). \ari{This is a great diagram. I think "form of `pairs'" is too confusing as a description though....how about ``paired format''?}
    the distribution of the input text, the label space, and the use of input-label pairing as the format of the \demo.
}\label{fig:composition}
\end{figure}
\begin{figure*}
\centering \scriptsize %\footnotesize
\setlength{\tabcolsep}{0.2em}
\begin{tabular}{llcccc}
    \makecell[l]{
    \resizebox{1.5\columnwidth}{!}{\includegraphics[trim={5cm 0cm 5cm 0cm},clip]{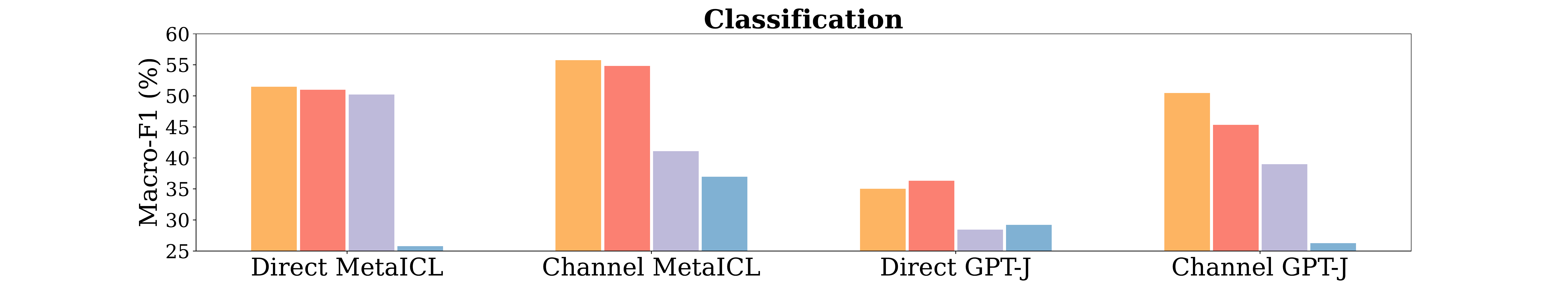}} \\
    \resizebox{1.5\columnwidth}{!}{\includegraphics[trim={5cm 0cm 5cm 0cm},clip]{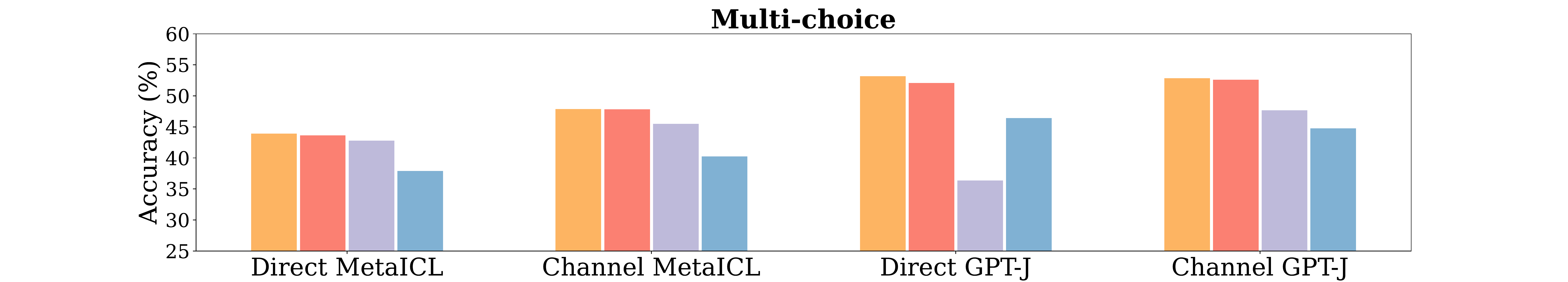}}}
    &
    \makecell[l]{
        \\
        \goldbox\ Gold labels \\ 
        \redbox\ Random labels \\ 
        \purplebox\ OOD + Random labels \\ 
        \bluebox\ No \demo \\ \\
        ~~{\em F: Format} \\
        ~~{\em L: Label space} \\
        ~~{\em I: Input distribution} \\
        ~~{\em M: Input-Label Mapping} \\
    } &
    \makecell[l]{
        {\em F} \\ \cmark \\ \cmark \\ \cmark \\ \xmark \\ \\  \\ \\ \\ \\
    } & 
    \makecell[l]{
        {\em L} \\ \cmark \\ \cmark \\ \cmark \\ \xmark \\ \\  \\ \\ \\ \\
    } & 
    \makecell[l]{
        {\em I} \\ \cmark \\ \cmark \\ \xmark \\ \xmark \\ \\  \\ \\ \\ \\
    } & 
    \makecell[l]{
        {\em M} \\ \cmark \\ \xmark \\ \xmark \\ \xmark \\ \\  \\ \\ \\ \\
    }
    \end{tabular}%\vspace{-.5em}
\caption{
    Impact of the distribution of the inputs.
    Evaluated in classification (top) and multi-choice (bottom).
    The impact of the distribution of the input text can be measured by comparing \redbox\ and \purplebox. 
    The gap is substantial, with an exception in Direct MetaICL
    (discussion in Section~\ref{subsec:abl_input}).
}\label{fig:abl_input}
\end{figure*}

We identify four aspects of the \demo\ $(x_1, y_1)...(x_k, y_k)$ that potentially provide learning signal (depicted in Figure~\ref{fig:composition}).
%(1) the example inputs, (2) the example labels, and (3) the fact that it consists of input-label ``pairs''
\begin{enumerate}[itemsep=0em]
    \item \textbf{The input-label mapping}, i.e., whether each input $x_i$ is paired with a correct label $y_i$.
    \item \textbf{The distribution of the input text}, i.e., the underlying distribution that $x_1...x_k$ are from.
    %the distribution of the text covered by $x_1...x_k$.
    \item \textbf{The label space}, i.e., the space covered by $y_1...y_k$.
    \item \textbf{The format}---specifically, the use of input-label pairing as the format.%\footnote{There are many aspects of the format, such as the separation between the input and the output, or different arrangements of the \demo. We only focus on the aspect of {\em using of input-label pairing} (instead of using inputs only or outputs only) and leave other aspects as future work.}
\end{enumerate}

%We hypothesize that components in the \demo\ that potentially lead to performance gains, aside the input-label mapping, are: (1) the input examples, (2) the label examples and (3) the input-label format.
%As we did in Section~\ref{sec:main} for the input-label mapping,
As Section~\ref{sec:main} does for the input-label mapping,
we design a series of variants of the \demo\ that quantify the impact of each aspect in isolation (Section~\ref{subsec:abl_input}--\ref{subsec:abl_format}). % (Table~\ref{tab:abl-methods}).
%We carefully design each variant to preserve the overall format of the \demo\ as much as possible.
We then additionally discuss the trend of the models meta-trained with an in-context learning objective (Section~\ref{subsec:abl_metaicl}).
For all experiments, models are evaluated on five classification and four multi-choice datasets as in Section~\ref{subsec:abl_main}.
See Appendix~\ref{app:exp-details} and Table~\ref{tab:example_demons} for implementation details and example \demo, respectively.
%Implementation details are reported in Appendix~\ref{app:exp-details}; Table~\ref{tab:example_demons} in the Appendix provides example \demo\ for each variant.

\begin{figure*}[!t]
\centering \scriptsize %\footnotesize
\setlength{\tabcolsep}{0.2em}
\begin{tabular}{llcccc}
    \makecell[l]{
    \resizebox{1.5\columnwidth}{!}{\includegraphics[trim={5cm 0cm 5cm 0cm},clip]{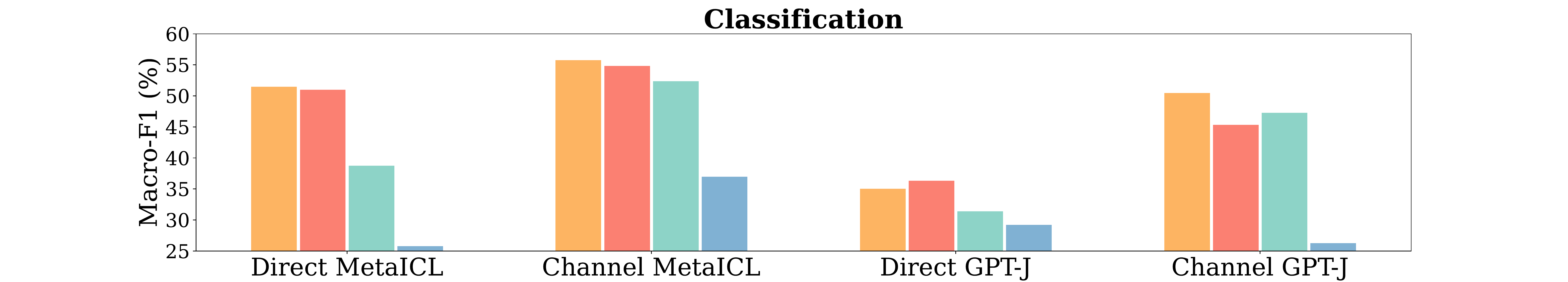}} \\
    \resizebox{1.5\columnwidth}{!}{\includegraphics[trim={5cm 0cm 5cm 0cm},clip]{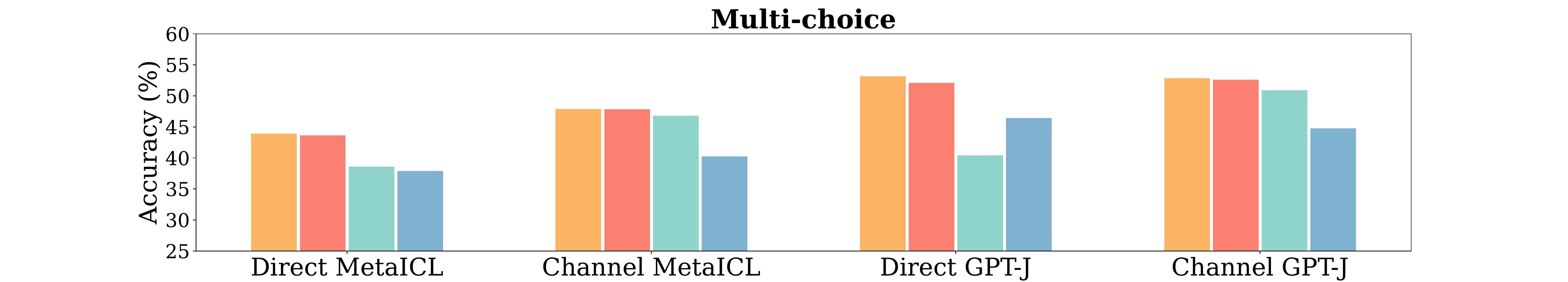}}}
    &
    \makecell[l]{
        \\
        \goldbox\ Gold labels \\ 
        \redbox\ Random labels \\ 
        \greenbox\ Random English words \\ 
        \bluebox\ No \demo \\ \\
        ~~{\em F: Format} \\
        ~~{\em L: Label space} \\
        ~~{\em I: Input distribution} \\
        ~~{\em M: Input-Label Mapping} \\
    } &
    \makecell[l]{
        {\em F} \\ \cmark \\ \cmark \\ \cmark \\ \xmark \\ \\  \\ \\ \\ \\
    } & 
    \makecell[l]{
        {\em L} \\ \cmark \\ \cmark \\ \xmark \\ \xmark \\ \\  \\ \\ \\ \\
    } & 
    \makecell[l]{
        {\em I} \\ \cmark \\ \cmark \\ \cmark \\ \xmark \\ \\  \\ \\ \\ \\
    } & 
    \makecell[l]{
        {\em M} \\ \cmark \\ \xmark \\ \xmark \\ \xmark \\ \\  \\ \\ \\ \\
    }
    \end{tabular}%\vspace{-.5em}
\caption{
    Impact of the label space.
    Evaluated in classification (top) and multi-choice (bottom).
    The impact of the label space can be measured by comparing \redbox\ and \greenbox. The gap is significant in the direct models but not in the channel models (discussion in Section~\ref{subsec:abl_label}).
}\label{fig:abl_label}
\end{figure*}

\subsection{Impact of the distribution of the input text}\label{subsec:abl_input}

%\paragraph{Methods.}
We experiment with \textbf{OOD \demo} which include out-of-distribution (OOD) text instead of the inputs from unlabeled training data. Specifically, a set of $k$ sentences $\{x_{i,\mathrm{rand}}\}_{i=1}^k$ are randomly sampled from an external corpus, and replace $x_1...x_k$ in the \demo. %Each sentence is paired with either a random label, a constant label or no labels, as we did with in-domain unlabeled data.
This variant assesses the impact of the distribution of the input text, while keeping the label space and the format of the \demo.
%conditioning on example inputs from the in-domain training data.

%\vspace{-.1em}
\paragraph{Results.}
Figure~\ref{fig:abl_input} shows that using out-of-distribution inputs instead of the inputs from the training data significantly drops the performance when Channel MetaICL, Direct GPT-J or Channel GPT-J are used, both in classification and multi-choice, by 3--16\% in absolute.
In the case of Direct GPT-J in multi-choice, it is even significantly worse than no \demo.
Direct MetaICL is an exception, which we think is the effect of meta-training (discussion in Section~\ref{subsec:abl_metaicl}).
%Using OOD inputs instead of inputs from the training data significantly drops the performance.\footnote{With an exception of Direct MetaICL, which we explain in the later of the section. \ari{I would just make this part of the running text}} In particular, in Direct GPT-J, using OOD inputs with random labels is worse than no-\demo.

This suggests that in-distribution inputs in the \demo\ substantially contribute to performance gains.
%We hypothesize that
This is likely because conditioning on the in-distribution text makes the task closer to language modeling, since the LM always conditioned on the in-distribution text during training.
%Nonetheless, it is worth noting that conditioning on the example inputs is not the {\em only} factor that leads to performance gains: the \demo\ with the example inputs and no labels does not substantially improve the performance over no-\demo. \ari{It would be worth pointing gout that this changes the format, and thus the ``trigger'' for producing an answer. Also maybe we want color-coded blocks to indicate the only-inputs to no-demos contrast?}

\subsection{Impact of the label space}\label{subsec:abl_label}

We also experiment with \textbf{\demo\ w/ random English words} that use random English words as labels for all $k$ pairs.
%replaces all labels in the \demo\ to the constant text, \texttt{answer}.
Specifically, we sample a random subset of English words $\mathcal{C}_\mathrm{rand}$ where $|\mathcal{C}_\mathrm{rand}|=|\mathcal{C}|$, and randomly pair $\tilde{y}_i \in \mathcal{C}_\mathrm{rand}$ with $x_i$.
This variant assesses the impact of the label space, while keeping the distribution of the input text and the format of the \demo.

\begin{figure*}[!t]
\centering \scriptsize %\footnotesize
\setlength{\tabcolsep}{0.2em}
\begin{tabular}{llcccc}
    \makecell[l]{
    \resizebox{1.5\columnwidth}{!}{\includegraphics[trim={5cm 0cm 5cm 0cm},clip]{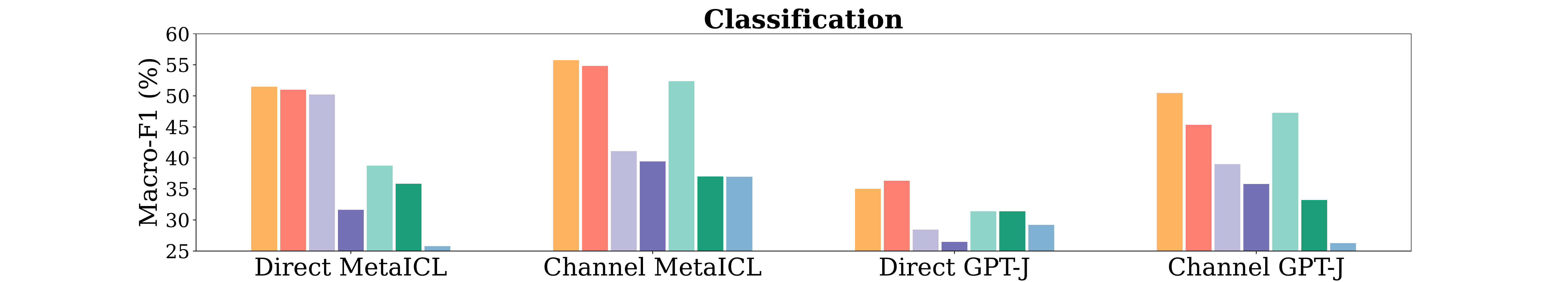}} \\
    \resizebox{1.5\columnwidth}{!}{\includegraphics[trim={5cm 0cm 5cm 0cm},clip]{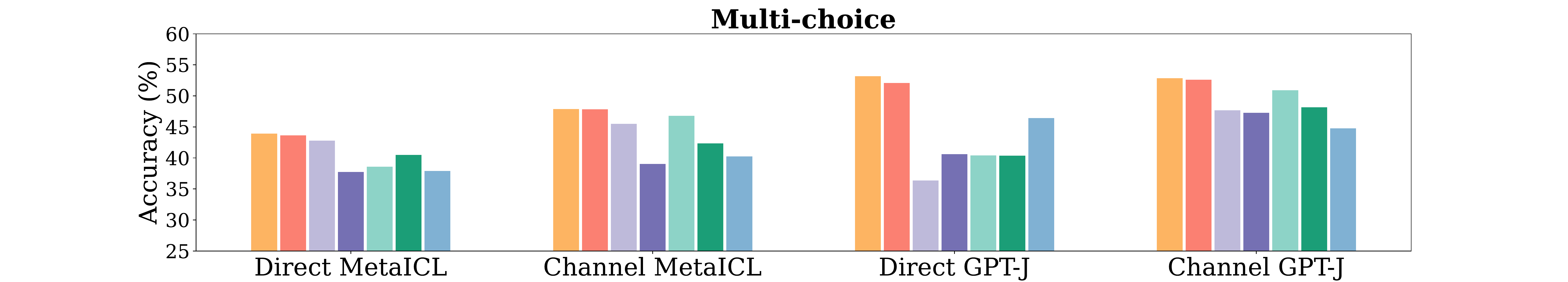}}}
    &
    \makecell[l]{
        \\
        \goldbox\ Gold labels \\ 
        \redbox\ Random labels \\ 
        \purplebox\ OOD + Random labels \\ 
        \darkpurplebox\ Random labels only \\  
        \greenbox\ Random English words \\ 
        \darkgreenbox\ No labels \\
        \bluebox\ No \demo \\ \\
        ~~{\em F: Format} \\
        ~~{\em L: Label space} \\
        ~~{\em I: Input distribution} \\
        ~~{\em M: Input-Label Mapping} \\
    } &
    \makecell[l]{
        {\em F} \\ \cmark \\ \cmark \\ \cmark \\ \xmark \\ \cmark \\ \xmark \\ \xmark \\ \\ \\ \\ \\ \\
    } & 
    \makecell[l]{
        {\em L} \\ \cmark \\ \cmark \\ \cmark \\ \cmark \\ \xmark \\ \xmark \\ \xmark \\ \\ \\ \\ \\ \\
    } & 
    \makecell[l]{
        {\em I} \\ \cmark \\ \cmark \\ \xmark \\ \xmark \\ \cmark \\ \cmark \\ \xmark \\ \\ \\ \\ \\ \\
    } & 
    \makecell[l]{
        {\em M} \\ \cmark \\ \xmark \\ \xmark \\ \xmark \\ \xmark \\ \xmark \\ \xmark \\ \\ \\ \\ \\ \\
    }
    \end{tabular}%\vspace{-.5em}
\caption{
    Impact of the format, i.e., the use of the input-label pairs.
    Evaluated in classification (top) and multi-choice (bottom).
    %The impact of the format can be examined via \purplebox\ vs. \darkpurplebox\ or \greenbox\ vs. \darkgreenbox.
    Variants of \demo\ without keeping the format (\darkpurplebox\ and \darkgreenbox) are overall not better than no \demo\ (\bluebox).
    Keeping the format is especially significant when it is possible to achieve substantial gains with the label space but without the inputs (\purplebox\ vs. \darkpurplebox\ in Direct MetaICL), or with the input distribution but without the labels (\greenbox\ vs. \darkgreenbox\ in Channel MetaICL and Channel GPT-J).
    More discussion in Section~\ref{subsec:abl_format}.
}\label{fig:abl_format}
\end{figure*}

%\vspace{-.1em}
\paragraph{Results.}
Based on Figure~\ref{fig:abl_label}, direct models and channel models exhibit different patterns.
With direct models, the performance gap between using random labels within the label space and using random English words is significant, ranging between 5--16\% absolute. This indicates that conditioning on the label space significantly contributes to performance gains.
This is true even for multi-choice tasks where there is no fixed set of labels---we hypothesize that multi-choice tasks still do have a particular distribution of the choices (e.g., objects like ``Bolts'' or ``Screws'' in the OpenBookQA dataset) that the model uses.

On the other hand, removing the output space does not lead to significant drop in the channel models: there is 0--2\% drop in absolute, or sometimes even an increase. % (with GPT-J in classification).
We hypothesize that this is because the channel models only condition on the labels, and thus are not benefiting from knowing the label space. This is in contrast to direct models which must {\em generate} the correct labels.
%\sewon{I'm not sure if this explanation is enough.}

\subsection{Impact of input-label pairing}\label{subsec:abl_format}

Section~\ref{subsec:abl_input} and \ref{subsec:abl_label} focus on variants which keep the format of the \demo\ as much as possible.
This section explores variants that change the format.
While there are many aspects of the format, we make minimal modifications to remove the pairings of inputs to labels.
%Given that the impact of the distributions of the input and the output space is significant,
%e.g., keep the input text only without labels, or keep the labels only without inputs.
Specifically, we evaluate \textbf{\demo\ with no labels} where the LM is conditioned on the concatenation of $x_1...x_k$, and \textbf{\demo\ with labels only} where the LM is conditioned on the concatenation of $y_1...y_k$. %Intuitively, the impact of the format can be measured by comparisons between OOD \demo\ and \demo\ with no labels, and between \demo\ with random English words and \demo\ with labels only.
These ablations provide the no-format counterparts of the `\demo\ with random English words' and `\demo\ with OOD inputs', respectively.

%\vspace{-.1em}
\paragraph{Results.}
Based on Figure~\ref{fig:abl_format}, 
removing the format is close to or worse than no \demo, indicating the importance of the format.
%Keeping the format is especially significant when its counterpart that keeps the format yields competitive performance, e.g., Direct MetaICL with the label space, or channel models with the input distribution.
This is likely because conditioning on a sequence of input-label pairs triggers the model to mimic the overall format and complete the new example as expected when the test input is given.

More interestingly, keeping the format plays a significant role in retaining a large portion of performance gains by only using the inputs or only using the labels.
For instance, with Direct MetaICL,
%which is relatively insensitive to out-of-distribution inputs,
it is possible to retain 95\% and 82\% of improvements from in-context learning (\demo\ with gold labels) by simply sampling random sentences from a corpus and randomly pairing them with the label set (\purplebox\ in Figure~\ref{fig:abl_format}) in classification and multi-choice, respectively.
Similarly, with the channel models,
%which are relatively insensitive to using labels outside of the label set,
it is possible to retain 82\%, 87\%, 86\% and 75\% of improvements from in-context learning by simply pairing each input from the unlabeled training data with a random English word (\greenbox\ in Figure~\ref{fig:abl_format}) in MetaICL classification, GPT-J classification, MetaICL multi-choice and GPT-J multi-choice, respectively.
For all of these cases, removing inputs instead of using OOD inputs, or removing labels instead of using random English words is significantly worse, indicating that \textbf{keeping the format of the input-label pairs is key}.

\subsection{Impact of meta-training}\label{subsec:abl_metaicl}

Different from other models, MetaICL is trained with an in-context learning objective, in line with recent work that uses multi-task training on a large collection of supervised datasets (called meta-training) 
for generalization to new tasks~\citep{aghajanyan2021muppet,khashabi2020unifiedqa,wei2022finetuned,sanh2022multitask}.
We aim to better understand the role of this meta-training in relation with our findings by closely examining the result of MetaICL.
In particular, we observe that the patterns we see so far are significantly more evident with MetaICL than with other models. For instance, the \gt\ input-label mapping matters even less, and keeping the format of the \demo\ matters even more.
There is nearly zero influence of the input-label mapping and the input distribution in Direct MetaICL, and the input-label mapping and the output space in Channel MetaICL.

Based on this observation, we hypothesize that \textbf{meta-training encourages the model to exclusively exploit simpler aspects of the \demo\ and to ignore others}.
%Specifically, Direct MetaICL mainly benefits from the format and the label space, while getting nearly zero improvements from the input-label mapping and the input distribution; Channel MetaICL, similarly, mainly benefits from the format and in-distribution inputs and ignores others.
This is based on our intuition that (1) the input-label mapping is likely harder to exploit, (2) the format is likely easier to exploit, and (3) the space of the text that the model is trained to generate is likely easier to exploit than the space of the text that the model conditions on.\footnote{
That is, the direct model exploits the label space better than the input distribution, and the channel model exploits the input distribution better than the label space.
}
%the direct model is likely to exploit the label space more easily while the channel model is likely to exploit the input distribution more easily.\footnote{This is because the direct model and the channel model are trained to maximize the likelihood of the label and the likelihood of the input, respectively.}

%Note that this is in line with observations from Section~\ref{sec:main} that there is less performance drop from removing the groundtruth input-label mapping in MetaICL than other LMs that are not finetuned.

%\vspace{.4em}
%To summarize, except for the input-label mapping, all of the example inputs, the example labels and the form the input-label pairs are overall important for the task performance: \textbf{all models require at least two out of three to achieve meaningful improvements over no-demonstration}.
%Although how much each component exactly matters varies across different types of models, results of the best models for each task indicate that the gains mostly come from both the example inputs and outputs, and also from the form of the input-label pairs in the case of classification.

%\section{Zero-shot through Demonstrations}\label{sec:silver}\input{sections/06-silver-zero-shot}
\section{Discussion \& Conclusion}\label{sec:discuss}In this paper, we study the role of the \demo\ with respect to the success of in-context learning.
We find that the \gt\ input-label mapping in the \demo\ matters significantly less than one might think---replacing gold labels with random labels in the \demo\ only marginally lowers the performance.
We then identify a series of aspects in the \demo\ and examine which aspect actually contributes to performance gains.
%While the results depend on the type of models and tasks, we find an overall trend in (1) significant impact of the distribution of the input and the label space, (2) significant impact of the format, e.g., it is possible to retain up to 95\% of gains from in-context learning by keeping the format and using either the inputs only or the labels only.
Results reveal that (1) gains are mainly coming from {\em independent} specification of the input space and the label space, (2) the models can still retain up to 95\% of performance gains by using either the inputs only or the label set only if the right format is used, and (3) meta-training with an in-context learning objective magnifies these trends.
Together, our findings lead to a set of broader indications about in-context learning, as well as avenues for future work.

\vspace{-.1em}
\paragraph{Does the model {\em learn} at test time?}
If we take a strict definition of learning: capturing 
%the task semantics
the input-label correspondence
given in the training data, then our findings suggest that LMs do not learn new tasks at test time.
Our analysis shows that the model may ignore the task defined by the \demo\ and instead use prior from pretraining.
%Our experiments in Section~\ref{subsec:abl_main} show that when the task is defined by the \demo\ to predict `negative' to a positive review and `positive' to a negative review, the model still predicts `positive' and `negative' to positive and negative reviews, respectively.

However, {\em learning} a new task can be interpreted more broadly: it may include adapting to specific input and label distributions and the format suggested by the \demo, and ultimately getting to make a prediction more accurately. With this definition of learning, the model {\em does} learn the task from the \demo.
Our experiments indicate that
the model {\em does} make use of aspects of the \demo\ and achieve performance gains.
%the \demo\ specify the distributions of inputs and labels and how they should be formatted, and the models make use of them to achieve performance gains.
%The \demo\ also removes the need of template engineering, or in other words, is the best form of the template.\footnote{Additional experiments in Appendix~\ref{app:template} show that minimal templates and engineered templates have little difference in performance, both significantly better than zero-shot with engineered templates.}

\vspace{-.1em}
\paragraph{Capacity of LMs.}
The model performs a downstream task without relying on the input-label correspondence from the \demo. This suggests that the model has learned the (implicit notion of) input-label correspondence from the language modeling objective alone, e.g., associating a positive review with the word `positive'.
This is in line with \citet{reynolds2021prompt} who claim that the \demo\ are for {\em task location} and the intrinsic ability to perform the task is obtained at pretraining time.\footnote{However, while \citet{reynolds2021prompt} claims that the \demo\ are thus unnecessary, we think using the \demo\ is actually the most unambiguous and the easiest way to prompt the model to perform a task.}

On one hand, this suggests that the language modeling objective has led to great zero-shot {\em capacity}, even if it is not always evident from the naive zero-shot {\em accuracy}.
On the other hand, this suggests that in-context learning %is unlikely to work on a task whose input-label correspondence is not already captured in the LM, e.g., when the task semantics are not close enough to language modeling.
may not work on a task whose input-label correspondence is not already captured in the LM.
%\ifshorten{}\else{
This leads to the research question of how to make progress in NLP problems that in-context learning does not solve: whether we need a better way of extracting the input-label mappings that are already stored in the LM, a better variant of the LM objective that learns a wider range of task semantics, or explicit supervision through fine-tuning on the labeled data.
%}\fi

%\vspace{-.1em}
%\paragraph{Connection to \citet{reynolds2021prompt}.}
%\citet{reynolds2021prompt} claim that the \demo\ are for {\em task location} and the intrinsic ability to perform the task has been obtained at pretraining time, which is in line with our observations.
%\citet{reynolds2021prompt} also claim that the model does not learn at test time, and the \demo\ are largely unnecessary.
%We do not necessarily agree with these claims: (1) {\em learning} can be interpreted more broadly as we already discussed, and (2) using the \demo\ is actually the clearest and the easiest way to prompt the model to perform a task.
%However, while \citet{reynolds2021prompt} claims that the \demo\ are thus unnecessary, we think using the \demo\ is actually the most unambiguous and the easiest way to prompt the model to perform a task.
%\ari{I disagree with this last part! The point seems to be that the model needs to be told what to do, and demonstrations are the easiest way of telling it.}\sewon{OK, now I made is super clear to what extent we agree with R\&M and what we disagree.}

\vspace{-.1em}
\paragraph{Connection to instruction-following models.}
Prior work has found it promising to train the model that reads the natural language description of the task (called instructions) and performs a new task at inference~\citep{mishra2021cross,efrat2020turking,wei2022finetuned,sanh2022multitask}. We think the \demo\ and instructions largely have the same role to LMs, and hypothesize that our findings hold for instruction-following models:
the instructions prompt the model to recover the capacity it already has, but do not supervise the model to learn novel task semantics.
This has been partially verified by \citet{webson2022prompt} who showed that the model performance does not degrade much with irrelevant or misleading instructions.
We leave more analysis on instruction-following
models for future work.

\vspace{-.1em}
\paragraph{Significantly improved zero-shot performance.}
One of our key findings is that it is possible to achieve nearly $k$-shot performance without using any labeled data, by simply pairing each unlabeled input with a random label and using it as the \demo.
This means our zero-shot baseline level is significantly higher than previously thought.\footnote{We take the perspective that using the unlabeled training data is permitted~\citep{kodirov2015unsupervised,wang2019survey,schick2021s}.}
%Gains from the \demo\ with random labels over the previous zero-shot method (no \demo) are up to 20\% absolute in classification and up to 15\% absolute in multi-choice tasks.
Future work can further improve the zero-shot performance with relaxed assumptions in access to the unlabeled training data.

%\mikel{One question that the paper doesn't address and could be interesting as future work is to understand if gradient-based few-shot methods (PET and similar) have a similar behavior. I would expect that these methods will be better at learning the input-label mapping, which would question how practical in-context learning is.}

\section*{Limitation}
%\updated{
    \paragraph{Effect of types of tasks and datasets.}
This paper focuses on the tasks from established NLP benchmarks that have {\em real} natural language inputs. Synthetic tasks with more limited inputs may actually use the \gt\ labels more, as observed by
    \citet{rong2021extrapolating}.
%}

We report macro-level analysis by examining the average performance over multiple NLP datasets, but different datasets may behave differently. Appendix~\ref{app:task-breakdown} discusses this aspect, including findings that there are larger gaps between using the \gt\ labels and using the random labels in some dataset-model pairs (e.g., in the most extreme case, nearly 14\% absolute on the financial\_phrasebank dataset with GPT-J).
Since the first version of our paper, \citet{kim2022ground} showed that using negated labels substantially lowers the performance in classification.\footnote{Note that \citet{kim2022ground} estimate the random label performance by interpolating with the performance using negated labels, while our paper samples the random labels at uniform.} We believe it is important to understand to what extend the model needs the \gt\ labels to successfully perform in-context learning.

\vspace{-.1em}
\paragraph{Extensions to generation.}
Our experiments are limited to classification and multi-choice tasks.
We hypothesize that ground truth output may not be necessary for in-context learning in the open-set tasks such as generation, but leave this to future work.
Extending of our experiments to such tasks is not trivial, because it requires a variation of the output which has incorrect input-output correspondence while keeping the correct output distribution (which is important based on our analysis in Section~\ref{sec:abl}).

Since the first version of our paper, \citet{madaan2022text} conducted a similar analysis with the chain of thought prompting~\citep{wei2022chain} which generates a rationale to perform complex tasks such as math problems.
\citet{madaan2022text} show that, while simply using a random rationale in the \demo\ (e.g., pairing with a rationale from a different example) significantly degrades the performance, other types of counterfactual rationales (e.g., wrong equations) do not degrade the performance as much as we thought.
We refer to \citet{madaan2022text} for more discussions on what aspects of the rationale matter or do not matter.

\section*{Acknowledgements}
We thank Gabriel Ilharco, Julian Michael, Ofir Press, UW NLP members and anonymous reviewers for their comments in the paper.
This research was supported by NSF IIS-2044660, ONR N00014-18-1-2826, a Sloan fellowship and gifts from AI2.

%\bibliographystyle{acl_natbib}
\bibliography{datasets,acl}

\clearpage
\appendix
\section{Full Datasets}\label{app:datasets}

We include 26 datasets as follows:
financial\_phrasebank~\citep{financial-phrasebank}, poem\_sentiment~\citep{sheng-uthus-2020-investigating}, medical\_questions\_pairs~\citep{medical-qqp}, glue-mrpc~\citep{dolan-brockett-2005-automatically}, glue-wnli~\citep{levesque2012winograd}, climate\_fever~\citep{Diggelmann2020CLIMATEFEVERAD}, glue-rte~\citep{dagan2005pascal, bar2006second,giampiccolo2007third, bentivogli2009fifth}, superglue-cb~\citep{Marneffe_Simons_Tonhauser_2019}, sick~\citep{marelli-etal-2014-sick} , hate\_speech18~\citep{gibert2018hate}, ethos-national\_origin~\citep{Mollas2020ETHOSAO}, ethos-race~\citep{Mollas2020ETHOSAO}, ethos-religion~\citep{Mollas2020ETHOSAO}, tweet\_eval-hate~\citep{barbieri-etal-2020-tweeteval}, tweet\_eval-stance\_atheism~\citep{barbieri-etal-2020-tweeteval}, tweet\_eval-stance\_feminist~\citep{barbieri-etal-2020-tweeteval}, quarel~\citep{Tafjord_Clark_Gardner_Yih_Sabharwal_2019}, openbookqa~\citep{mihaylov-etal-2018-suit}, qasc~\citep{Khot_Clark_Guerquin_Jansen_Sabharwal_2020}, commonsense\_qa~\citep{talmor-etal-2019-commonsenseqa}, ai2\_arc~\citep{Clark2018ThinkYH}, codah~\citep{chen-etal-2019-codah}, superglue-copa~\citep{gordon-etal-2012-semeval}, dream~\citep{sun-etal-2019-dream}, quartz-with\_knowledge~\citep{tafjord-etal-2019-quartz}, quartz-no\_knowledge~\citep{tafjord-etal-2019-quartz}.
The choice of datasets is made following low-resource datasets in \citet{min2021metaicl}, with the exact same set of $k$-shot train data using 5 random seeds.
We use the HuggingFace version of the data~\citep{lhoest-etal-2021-datasets} and use the development data for evaluation, following \citet{ye2021crossfit}.
See Table~\ref{tab:data} for statistics.

\begin{table}[t]
    \centering \footnotesize
    \begin{tabular}{lrr}
        \toprule
            Dataset & \# Train & \# Eval \\
        \midrule
            \multicolumn{2}{l}{\em Task category: Sentiment analysis} \\
            financial\_phrasebank       & 1,811 & 453 \\
            poem\_sentiment             & 892 & 105 \\
        \midrule
            \multicolumn{2}{l}{\em Task category: Paraphrase detection} \\
            medical\_questions\_pairs   & 2,438 & 610 \\
            glue-mrpc                   & 3,668 & 408 \\
        \midrule
            \multicolumn{2}{l}{\em Task category: Natural language inference} \\
            glue-wnli                   & 635 & 71 \\
            climate\_fever              & 1,228 & 307 \\
            glue-rte                    & 2,490 & 277 \\
            superglue-cb                & 250 & 56 \\
            sick                        & 4,439 & 495 \\
        \midrule
            \multicolumn{2}{l}{\em Task category: Hate speech detection} \\
            hate\_speech18              & 8,562 & 2,141 \\
            ethos-national\_origin      & 346 & 87 \\
            ethos-race                  & 346 & 87 \\
            ethos-religion              & 346 & 87 \\
            tweet\_eval-hate            & 8,993 & 999 \\
            tweet\_eval-stance\_atheism & 461 & 52\\
            tweet\_eval-stance\_feminist & 597 & 67 \\
        \midrule
            \multicolumn{2}{l}{\em Task category: Question answering} \\
            quarel                      & 1,941 & 278 \\
            openbookqa                  & 4,957 & 500 \\
            qasc                        & 8,134 & 926 \\
            commonsense\_qa             & 9,741 & 1,221 \\
            ai2\_arc                    & 1,119 & 299 \\
        \midrule
            \multicolumn{2}{l}{\em Task category: Sentence completion} \\
            codah                       & 1665 & 556 \\
            superglue-copa              & 400 & 100 \\
            dream                       & 6116 & 2040\\
            quartz-with\_knowledge      & 2696 & 384\\
            quartz-no\_knowledge        & 2696 & 384 \\
        \bottomrule
    \end{tabular}\vspace{-.1em}
    \caption{26 datasets used for experiments, classified into 6 task categories. \# Train and \# Test indicate the number of training and test examples of the dataset.
    Note that \# train is based on the original training dataset but we use $k$ random samples for $k$-shot evaluation.
    }\label{tab:data}
\end{table}

\section{Experimental Details}\label{app:exp-details}

\paragraph{Example template}
We follow \citet{ye2021crossfit,min2021metaicl,logan2021cutting} in using the minimal format to transform the input to a sequence (e.g. a concatenation of multiple inputs) and using the label words from each dataset as it is.
We also explore manual templates taken from prior work~\citep{holtzman2021surface,zhao2021calibrate} as reported in Section~\ref{subsec:abl_main}, although we find that using these templates is not consistently better than using minimal templates.
We thus run main experiments with minimal templates. Example templates are provided in Table~\ref{tab:template-examples}.

\paragraph{Format of the \demo}
We follow the standard of each model for formatting the \demo, either from exploration in prior work or the example code provided in the official tutorial.
%\footnote{For instance, \href{https://github.com/pytorch/fairseq/tree/main/examples/moe_lm}{github.com/pytorch/fairseq/tree/main/examples/moe\_lm}}
For GPT-2, we separate the input and the label, and each demonstration example with a space. For MetaICL, GPT-J and GPT-3, we separate the input and the label with a newline (\texttt{\textbackslash{}n}), and each demonstration example with three newlines. For fairseq models, we use a newline to separate the input and the label as well as each demonstration example.

\begin{algorithm}[t]
%\captionsetup{font=10pt}
\caption{Forming the \demo\ with an accuracy of $a\%$.}\label{alg:abl_accuracy}
\begin{algorithmic}[1] \small
\Procedure{\textsc{FormDemons}}{$\{(x_i,y_i)\}_{i=1}^k, a$}
\State $D \gets [ ]$ ~~{\protect\color{gitgreen} {\textit{// demonstration to be formed}}} 
\State $n \gets k \times a/100$ {\protect\color{gitgreen}  ~~{\textit{// number of correct pairs}}} 
\State $\mathcal{G} \gets \mathrm{Sample}(\mathrm{Range}(1,k), n)$
\For{$i \in \mathrm{Range}(1,k)$}
    \If {$i \in \mathcal{G}$}
        ~~{\protect\color{gitgreen}  {\textit{// add correct pair}}} 
        \State $D\mathrm{.append}((x_i, y_i))$
    \Else
        ~~{\protect\color{gitgreen}  {\textit{// add incorrect pair}}} 
        \State $D\mathrm{.append}((x_i, \mathrm{Sample}(\mathcal{C} - y_i)))$
    \EndIf
\EndFor
\State \Return $D$
\EndProcedure
\end{algorithmic}
\end{algorithm}

\begin{figure*}
\centering \footnotesize
\resizebox{2.1\columnwidth}{!}{\includegraphics[trim={0 0.5cm 0 0.5cm},clip]{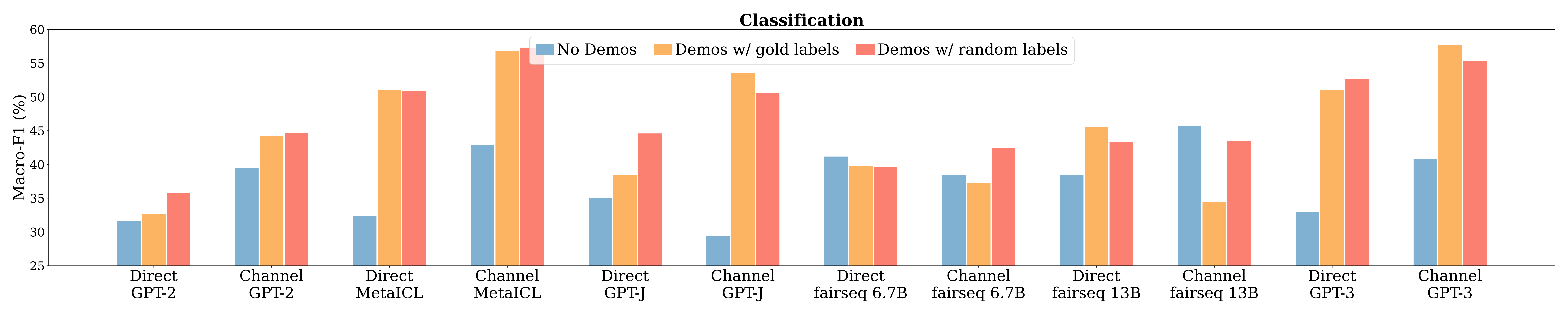}}
\resizebox{2.1\columnwidth}{!}{\includegraphics[trim={0 0.5cm 0 0.5cm},clip]{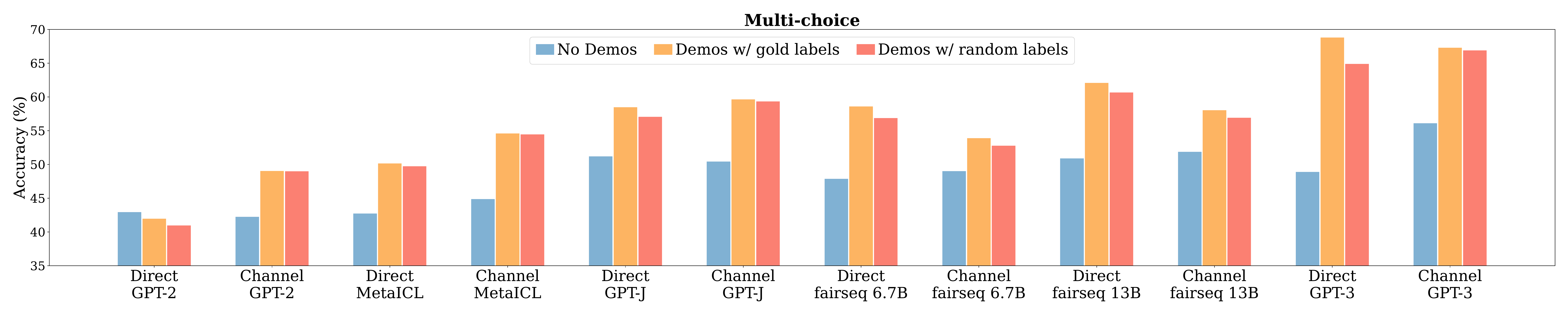}}
\caption{
    Results of No-demonstration, Gold demonstration and Random demonstration on 3 classification datasets (top) and 3 multi-choice datasets (bottom).
    Details in Section~\ref{subsec:main}.
    This figure is for providing numbers that are comparable across models---full results with more datasets are reported in Figure~\ref{fig:main-results}.
}\label{fig:app-main-results}
\end{figure*}

\paragraph{Details in variants of the \demo}

For ``\demo\ w/ $a$\% accurate labels'' ($0 \leq a \leq 100$), we use $k \times a/100$ correct pairs and $k \times (1-a/100)$ incorrect pairs in a random order, as described in Algorithm~\ref{alg:abl_accuracy}.
For ``OOD \demo'', we use CC-News~\citep{nagel2016cc} as an external corpus. We consider the length of the text during sampling, so that sampled sentences have similar length to the test input.
For ``\demo\ with random English words'', we use \href{https://pypi.org/project/english-words}{\nolinkurl{pypi.org/project/english-words}} for the set of English words, which consists of 61,569 words.

Table~\ref{tab:example_demons} provides a list of example \demo\ for each method used in Section~\ref{sec:abl}.

\section{More Experimental Results}\label{app:exp-results}
\subsection{Gold labels vs. random labels}
Figure~\ref{fig:app-main-results} shares the same interface as Figure~\ref{fig:main-results}, but all models are evaluated on 3 classification and 3 multi-choice datasets and are thus comparable to each other.

\subsection{Random labels from true distribution of labels \& Task breakdown}\label{app:task-breakdown}

In Section~\ref{sec:main}, random labels are sampled from the label space from a uniform distribution. We experiment with another variant of \demo\ in the classification tasks, where labels are randomly sampled from the true distribution of labels on the training data. This may have large impact if labels are far from uniform on the training data. Results indicate that performance drop from using gold labels is further reduced compared to using uniformly random labels: with Channel MetaICL, the gap is reduced from 1.9\% to 1.3\% absolute, and with Channel GPT-J, the gap is reduced from 5.0\% to 3.5\% absolute.

\begin{figure*}
\centering \footnotesize
\resizebox{2.1\columnwidth}{!}{\includegraphics[trim={0 0.5cm 0 0.5cm},clip]{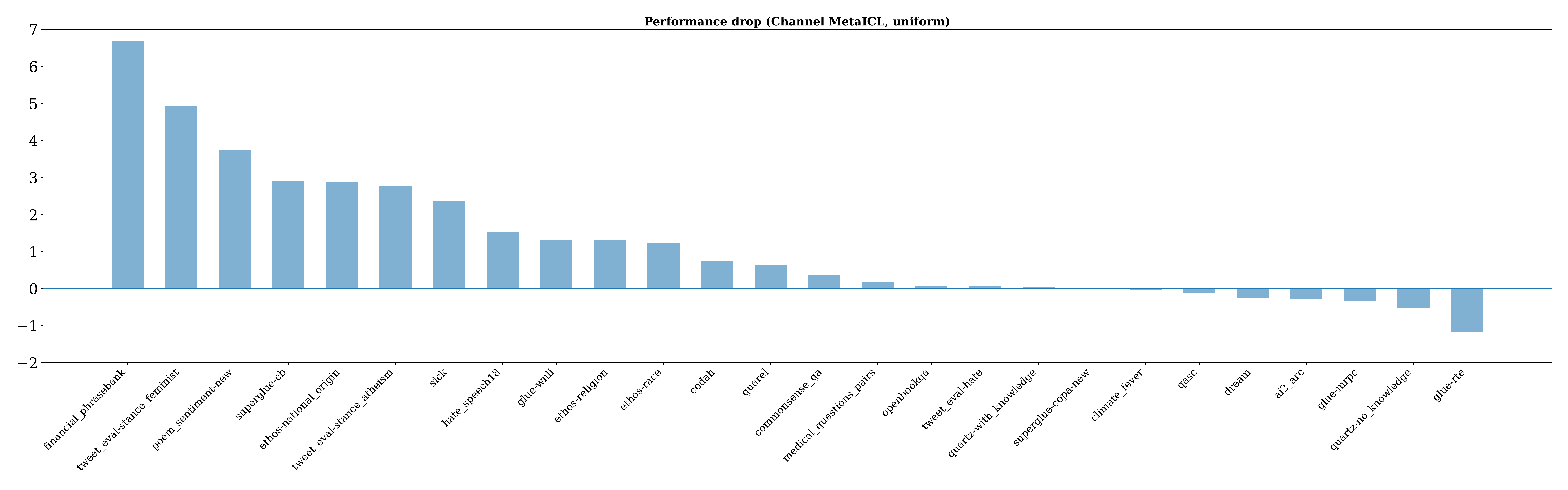}}
\resizebox{2.1\columnwidth}{!}{\includegraphics[trim={0 0.5cm 0 0.5cm},clip]{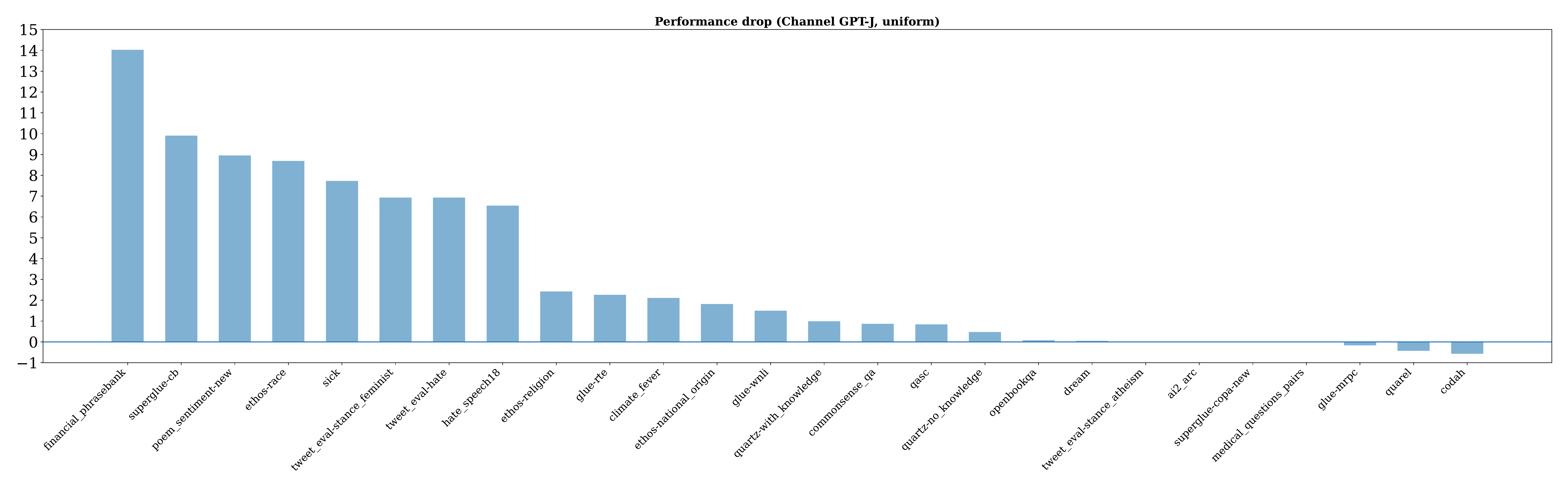}}
\resizebox{2.1\columnwidth}{!}{\includegraphics[trim={0 0.5cm 0 0.5cm},clip]{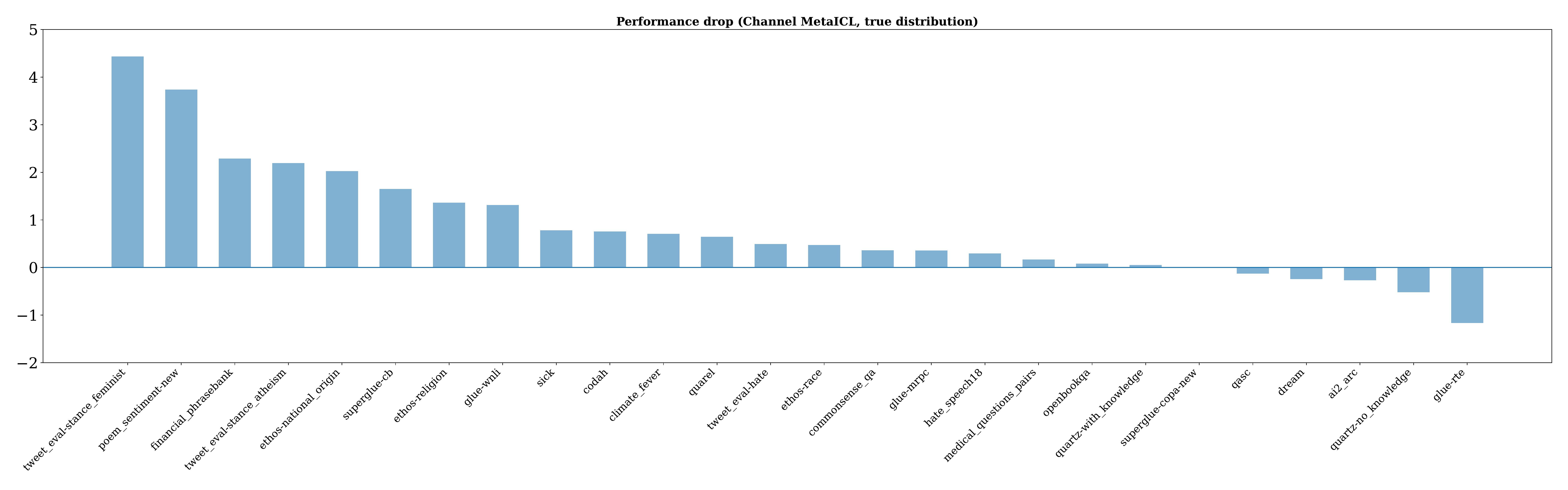}}
\resizebox{2.1\columnwidth}{!}{\includegraphics[trim={0 0.5cm 0 0.5cm},clip]{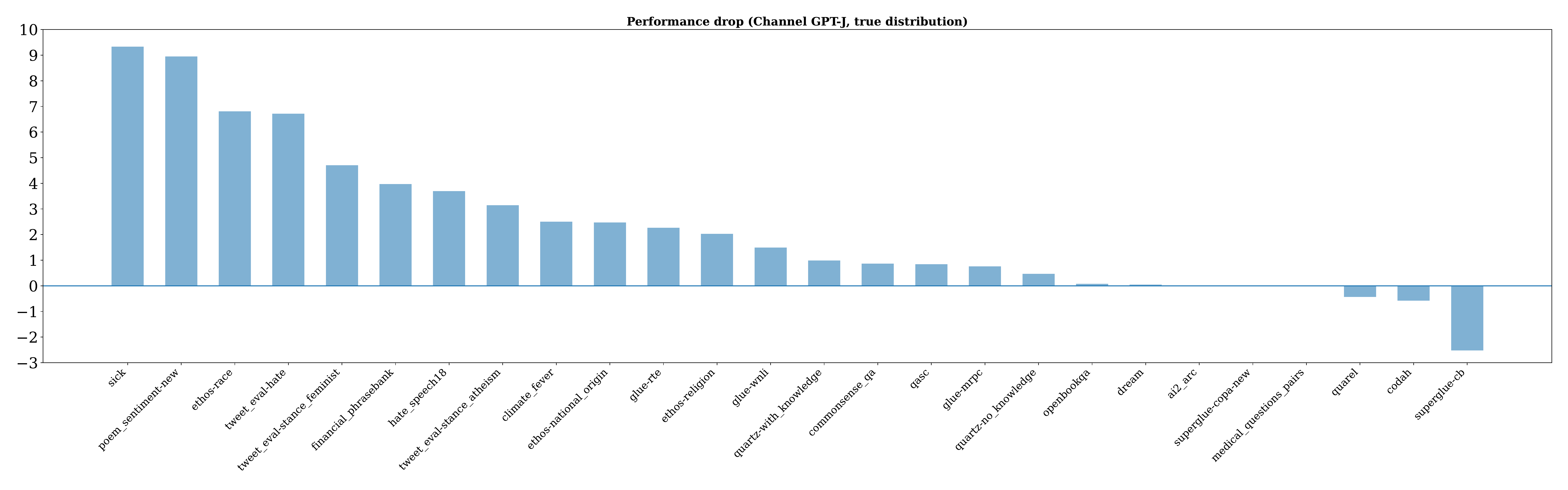}}
\caption{
    Performance gap from using the \demo\ with gold labels to using the \demo\ with random labels.
    Datasets are sorted in descending order.
    The top two figures use random labels that are sampled at uniform, with Channel MetaICL and Channel GPT-J, respectively.
    The bottom two figures use random labels that are sampled from a true distribution of labels on the training data, with Channel MetaICL and Channel GPT-J, respectively.
}\label{fig:task-breakdown}
\end{figure*}

Figure~\ref{fig:task-breakdown} shows performance gap between using gold labels and using random labels per dataset. We find that the trend that the gap is smaller than previously thought is consistant across most datasets. Nonetheless, there are a few outlier datasets where performance gap is non-negligible, such as financial\_phrasebank and a few hate speech detection datasets. Future work may investigate on which tasks the model makes more use of the correctly paired training data.

\subsection{More variants of the \demo}

We explored \textbf{\demo\ with a constant label} where all labels in the \demo\ are replaced with a constant text, ``\texttt{answer}''. 
Specifically, a prediction is made via $\mathrm{argmax}_{y \in \mathcal{C}}P(y|x_1,\texttt{answer}...x_k,\texttt{answer},x)$.
This can be viewed as another way to remove the impact of the label space while keeping the impact of the distribution of the input text.
However, results are consistently worse than the results of \demo\ with random English labels.
%, e.g., 36.0 vs. 38.7, 47.5 vs. 52.4, 29.3 vs. 31.4, 39.8 vs. 47.3
%40.5 vs. 38.6, 43.0 vs. 46.8, 41.2 vs. 40.4, 48.9 vs. 50.9
We think this is because constant labels actually change the format of the \demo, since they can be viewed as part of a separator between different demonstration examples.

We also explored \textbf{\demo\ with the test input} where all inputs in the \demo\ are replaced with the test input, each paired with a random label.
Specifically, a prediction is made via
$\mathrm{argmax}_{y \in \mathcal{C}}P(y|x,\tilde{y}_1...x,\tilde{y}_k,x)$, where $\tilde{y}_i$ ($1 \leq i \leq k$) is randomly sampled at uniform from $\mathcal{C}$.
This variant is seemingly a reasonable choice given that it satisfies the condition that the inputs in the \demo\ come from the same distribution as the test input (since they are identical), and using random labels is as good as using gold labels. 
Nonetheless, we find that this variant is significantly worse than most other methods with \demo. We think this is because using the constant input for all demonstration example significantly changes the format of the sequence, since the input can be viewed as part of a separator between different demonstration examples.

% 46.9 vs. 39.9 vs. 42.8
% 51.4 vs. 43.9 vs. 29.4
% 26.7 vs. 33.0 vs. 32.8
% 24.3 vs. 39.5 vs. 35.1

\definecolor{redtext}{HTML}{d53e4f}
\definecolor{bluetext}{HTML}{3288bd}
\newcommand{\redtext}[1]{\textcolor{redtext}{#1}}
\newcommand{\bluetext}[1]{\textcolor{bluetext}{#1}}
\newcommand{\purpletext}[1]{\textcolor{violet}{#1}}

\begin{table*}[t]
    \centering \footnotesize
    \setlength{\tabcolsep}{0.5em}
    \begin{tabular}{lll}
        \toprule
            Dataset & Type & Example\\
        \midrule
            \multirow{2}{*}{MRPC} & Minimal & \makecell[l]{
            \redtext{sentence 1: Cisco pared spending to compensate for sluggish sales . [SEP] sentence 2: In} \\ \redtext{response to sluggish sales , Cisco pared spending .} \textbackslash{n} \{\bluetext{equivalent}$\vert$\bluetext{not\_equivalent}\}
            } \\
            \cmidrule(lr){2-3}
            & Manual & \makecell[l]{
            \redtext{Cisco pared spending to compensate for sluggish sales .} \textbackslash{n} The question is: \redtext{In response to} \\ \redtext{sluggish sales , Cisco pared spending .} True or False? \textbackslash{n} The answer is:\{\bluetext{True}$\vert$\bluetext{False}\}} \\ 
        \midrule
            \multirow{2}{*}{RTE} & Minimal & \makecell[l]{\redtext{sentence 1: The girl was found in Drummondville. [SEP] sentence 2: Drummondville} \\ \redtext{contains the girl.}  \textbackslash{n}  \{\bluetext{entailment}$\vert$\bluetext{not\_entailment}\}} \\
            \cmidrule(lr){2-3}
            & Manual & \makecell[l]{\redtext{The girl was found in Drummondville.} \textbackslash{n} The question is: \redtext{Drummondville contains the} \\ \redtext{girl.} True or False? \textbackslash{n} The answer is:\{\bluetext{True}$\vert$\bluetext{False}\}} \\
        \midrule
            \multirow{2}{*}{Tweet\_eval-hate} & Minimal & \redtext{The Truth about \#Immigration} \textbackslash{n} \{\bluetext{hate}$\vert$\bluetext{non-hate}\} \\
            \cmidrule(lr){2-3}
            & Manual & Tweet: \redtext{The Truth about \#Immigration} \textbackslash{n} Sentiment: \{\bluetext{against}$\vert$\bluetext{favor}\}
            \\
        \midrule
            \multirow{2}{*}{SICK} & Minimal & \makecell[l]{\redtext{sentence 1: A man is screaming. [SEP] sentence 2: A man is scared.} \textbackslash{n} \\
            \{\bluetext{contradiction}$\vert$\bluetext{entailment}$\vert$\bluetext{neutral}\}} \\
            \cmidrule(lr){2-3}
            & Manual & \makecell[l]{\redtext{A man is screaming.} \textbackslash{n} The question is: \redtext{A man is scared}. True or False? \textbackslash{n} The answer is: \\ \{\bluetext{False}$\vert$\bluetext{True}$\vert$\bluetext{Not sure}\}} \\
        \midrule
            \multirow{2}{*}{poem-sentiment} & Minimal & \redtext{willis sneered:} \textbackslash{n} \{\bluetext{negative}$\vert$\bluetext{no\_impact}$\vert$\bluetext{positive}\} \\
            \cmidrule(lr){2-3}
            & Manual & \redtext{willis sneered:} \textbackslash{n} The sentiment is:  \{\bluetext{negative}$\vert$\bluetext{no\_impact}$\vert$\bluetext{positive}\}
            \\
        \midrule
            \multirow{2}{*}{OpenbookQA} & Minimal & \redtext{What creates a valley?} \textbackslash{n} \{\bluetext{feet}$\vert$\bluetext{rock}$\vert$\bluetext{water}$\vert$\bluetext{sand}\} \\
            \cmidrule(lr){2-3}
            & Manual & The question is: \redtext{What creates a valley?} \textbackslash{n} The answer is: \{\bluetext{feet}$\vert$\bluetext{rock}$\vert$\bluetext{water}$\vert$\bluetext{sand}\}
            \\
        \midrule
            \multirow{2}{*}{CommonsenseQA} & Minimal & \redtext{What blocks sunshine?} \textbackslash{n} \{\bluetext{summer}$\vert$\bluetext{park}$\vert$\bluetext{desktop}$\vert$\bluetext{sea}$\vert$\bluetext{moon}\} \\
            \cmidrule(lr){2-3}
            & Manual & The question is: \redtext{What blocks sunshine?} \textbackslash{n} The answer is: \{\bluetext{summer}$\vert$\bluetext{park}$\vert$\bluetext{desktop}$\vert$\bluetext{sea}$\vert$\bluetext{moon}\}
            \\
        \midrule
            \multirow{2}{*}{COPA} & Minimal & \redtext{Effect: I coughed.} \textbackslash{n} \{\bluetext{Cause: I inhaled smoke.}$\vert$\bluetext{Cause: I lowered my voice.}\} \\
            \cmidrule(lr){2-3}
            & Manual & \redtext{I coughed} because \{\bluetext{I inhaled smoke.}$\vert$\bluetext{I lowered my voice.}\}
            \\
        \midrule
            \multirow{2}{*}{ARC} & Minimal & \redtext{Which biome has the most vegetation?} \textbackslash{n} \{\bluetext{desert}$\vert$\bluetext{forest}$\vert$\bluetext{grassland}$\vert$\bluetext{tundra}\} \\
            \cmidrule(lr){2-3}
            & Manual & \makecell[l]{The question is: \redtext{Which biome has the most vegetation?} \textbackslash{n} The answer is: \{\bluetext{desert}$\vert$\bluetext{forest}$\vert$\\ \bluetext{grassland}$\vert$\bluetext{tundra}\}
            } \\
        \bottomrule
    \end{tabular}\vspace{-.1em}
    \caption{
        A list of minimal templates taken from \citet{ye2021crossfit,min2021metaicl} and manual templates taken from \citet{holtzman2021surface,zhao2021calibrate}.
        Details provided in Appendix~\ref{app:exp-details}.
        See Figure~\ref{fig:abl_template} for discussion in empirical results.
        %The inputs and labels are enclosed by the curly brackets.
        The input and the label are in the \redtext{red text} and in the \bluetext{blue text}, respectively.
        Note that $\vert$ is used to separate different options for the labels.
    }\label{tab:template-examples}
\end{table*}

\newcommand{\example}[4]{
    \multirow{3}{*}{\makecell[l]{{\em Demos} \\ {\em #1}}} &
    #2 \\
    & #3 \\
    & #4 \\
}
\newcommand{\oodexample}[4]{
    \multirow{3}{*}{\makecell[l]{{\em OOD Demos} \\ {\em #1}}} &
    #2 \\
    & #3 \\
    & #4 \\
}

\begin{table*}[t]
    \centering \footnotesize
    \setlength{\tabcolsep}{0.5em}
    \begin{tabular}{ll}
        \toprule
            \example{w/ gold labels}{
                ({\em Format} \cmark\ {\em Input distribution} \cmark\ {\em Label space} \cmark\ {\em Input-label mapping} \cmark)
            }{Circulation revenue has increased by 5\% in Finland and 4\% in Sweden in 2008. \textbackslash{}n positive}{Panostaja did not disclose the purchase price. \textbackslash{}n neutral}
        \midrule
            \example{w/ random labels}{
                ({\em Format} \cmark\ {\em Input distribution} \cmark\ {\em Label space} \cmark\ {\em Input-label mapping} \xmark)
            }{Circulation revenue has increased by 5\% in Finland and 4\% in Sweden in 2008. \textbackslash{}n \bluetext{neutral}}{Panostaja did not disclose the purchase price. \textbackslash{}n \bluetext{negative}}
        \midrule
            \oodexample{w/ random labels}{
                ({\em Format} \cmark\ {\em Input distribution} \xmark\ {\em Label space} \cmark\ {\em Input-label mapping} \xmark)
            }{\redtext{Colour-printed lithograph. Very good condition. Image size: 15 x 23 1/2 inches.} \textbackslash{}n \bluetext{neutral}}{\redtext{Many accompanying marketing claims of cannabis products are often well-meaning.} \textbackslash{}n \bluetext{negative}}
        \midrule
            \example{w/ random English words}{
                ({\em Format} \cmark\ {\em Input distribution} \cmark\ {\em Label space} \xmark\ {\em Input-label mapping} \xmark)
            }{Circulation revenue has increased by 5\% in Finland and 4\% in Sweden in 2008.
            \textbackslash{}n \purpletext{unanimity}}{Panostaja did not disclose the purchase price. \textbackslash{}n \purpletext{wave}}
        \midrule
        %    \example{w/ \texttt{answer}}{Circulation revenue has increased by 5\% in Finland and 4\% in Sweden in 2008.
        %    \textbackslash{}n answer}{Panostaja did not disclose the purchase price. \textbackslash{}n answer}
        %\midrule
            \example{w/o labels}{
                ({\em Format} \xmark\ {\em Input distribution} \cmark\ {\em Label space} \xmark\ {\em Input-label mapping} \xmark)
            }{Circulation revenue has increased by 5\% in Finland and 4\% in Sweden in 2008.}{Panostaja did not disclose the purchase price.}
        \midrule
            \example{labels only}{
                ({\em Format} \xmark\ {\em Input distribution} \xmark\ {\em Label space} \cmark\ {\em Input-label mapping} \xmark)
            }{\bluetext{positive}}{\bluetext{neutral}}
        %\midrule
        %    \oodexample{w/ \texttt{answer}}{Colour-printed lithograph. Very good condition. Image size: 15 x 23 1/2 inches. \textbackslash{}n answer}{Many accompanying marketing claims of cannabis products are often well-meaning. \textbackslash{}n answer}
        %\midrule
        %    \oodexample{w/o labels}{Colour-printed lithograph. Very good condition. Image size: 15 x 23 1/2 inches.}{Many accompanying marketing claims of cannabis products are often well-meaning.}
        \bottomrule
    \end{tabular}\vspace{-.1em}
    \caption{Example \demo\ when using methods in Section~\ref{sec:abl}.
    The financial\_phrasebank dataset with $\mathcal{C}=\{$``positive'', ``neutral'', ``negative''$\}$ is used.
    \redtext{Red text} indicates the text is sampled from an external corpus; \bluetext{blue text} indicates the labels are randomly sampled from the label set; \purpletext{purple text} indicates a random English word.
    %Only the first method uses labeled data.
    %First four methods use in-domain input text, and the rest three methods use random text sampled from CC-News.
    }\label{tab:example_demons}
\end{table*}

\end{document}